\documentclass[preprint,12pt,authoryear]{elsarticle}

\usepackage{amssymb}

\usepackage[utf8]{inputenc} 
\usepackage[T1]{fontenc}    
\usepackage{hyperref}       
\usepackage{url}            
\usepackage{booktabs}       
\usepackage{amsfonts}       
\usepackage{amsmath}        
\usepackage{nicefrac}       
\usepackage{microtype}      
\usepackage{xcolor}         

\hypersetup{
    colorlinks,
    linkcolor={blue!10!black},
    citecolor={blue!10!black},
    urlcolor={blue!10!black}
}

\usepackage{acro} 
\DeclareAcronym{ai}{
  short=AI,
  long=artificial intelligence,
}

\DeclareAcronym{mtnlg}{
  short=MT-NLG,
  long=Megatron-Turing natural language generation,
}

\DeclareAcronym{dnn}{
  short=DNN,
  long=deep neural network,
}

\DeclareAcronym{ocs}{
  short=OCS,
  long=optical circuit switched,
}

\DeclareAcronym{eps}{
  short=EPS,
  long=electronic packet switched,
}

\DeclareAcronym{rl}{
  short=RL,
  long=reinforcement learning,
}

\DeclareAcronym{gnn}{
  short=GNN,
  long=graph neural network,
}

\DeclareAcronym{jct}{
  short=JCT,
  long=job completion time,
}

\DeclareAcronym{mdp}{
    short=MDP,
    long=Markov decision process,
}

\DeclareAcronym{ppo}{
    short=PPO,
    long=proximal policy optimisation,
}

\DeclareAcronym{dag}{
    short=DAG,
    long=directed acyclic graph,
}

\DeclareAcronym{mpi}{
    short=MPI,
    long=message passing interface,
}

\DeclareAcronym{ml}{
    short=ML,
    long=machine learning,
}

\DeclareAcronym{co}{
    short=CO,
    long=combinatorial optimisation,
}

\DeclareAcronym{ilp}{
    short=ILP,
    long=integer linear programme,
}

\DeclareAcronym{dqn}{
    short=DQN,
    long=deep Q-learning,
}

\usepackage{siunitx} 
\usepackage{numprint} 

\usepackage[skip=2pt]{caption}

\usepackage{caption}
\newcommand{\secInLine}[1]{\textbf{#1}.}

\usepackage{rotating} 

\journal{Journal of Parallel and Distributed Computing}

\begin{document}

\begin{frontmatter}




\title{Partitioning Distributed Compute Jobs with Reinforcement Learning and Graph Neural Networks}


\author{Christopher W. F. Parsonson\corref{cor1}\fnref{label1}}
\author{Zacharaya Shabka\fnref{label1}}
\author{Alessandro Ottino\fnref{label1}}
\author{Georgios Zervas\fnref{label1}}

\cortext[cor1]{Corresponding author: zciccwf@ucl.ac.uk}

\fntext[label1]{UCL}




\begin{highlights}
\item Demonstrate that deciding \textit{how much} to partition distributed jobs is a key factor in determining overall system throughput.
\item Demonstrate that optimising for only the job completion time leads to high blocking rates and poor throughput in dynamic job arrival scenarios.
\item Introduce a new partitioning algorithm which leverages reinforcement learning, a graph neural network, and a novel formulation of the user-defined job completion time specification to automatically learn to partition jobs such that the blocking rate is minimised and user requirements are met.
\item Demonstrate the proposed algorithm out-performing baselines on a state-of-the-art optical network architecture running five real deep learning computation graphs.
\end{highlights}

\maketitle



\begin{abstract}

From natural language processing to genome sequencing, large-scale machine learning models are bringing advances to a broad range of fields. Many of these models are too large to be trained on a single machine, and instead must be distributed across multiple devices. This has motivated the research of new compute and network systems capable of handling such tasks. In particular, recent work has focused on developing management schemes which decide \textit{how} to allocate distributed resources such that some overall objective, such as minimising the job completion time (JCT), is optimised. However, such studies omit explicit consideration of \textit{how much} a job should be distributed, usually assuming that maximum distribution is desirable. In this work, we show that maximum parallelisation is sub-optimal in relation to user-critical metrics such as throughput and blocking rate. To address this, we propose PAC-ML (\underline{\textbf{p}}artitioning for \underline{\textbf{a}}synchronous \underline{\textbf{c}}omputing with \underline{\textbf{m}}achine \underline{\textbf{l}}earning). PAC-ML leverages a graph neural network and reinforcement learning to learn how much to partition computation graphs such that the number of jobs which meet arbitrary user-defined JCT requirements is maximised. In experiments with five real deep learning computation graphs on a recently proposed optical architecture across four user-defined JCT requirement distributions, we demonstrate PAC-ML achieving up to $56.2\%$ lower blocking rates in dynamic job arrival settings than the canonical maximum parallelisation strategy used by most prior works. 

\end{abstract}

\end{frontmatter}




\section{Introduction}
\label{sec:introduction}

The last decade has seen an exponential increase in the amount of compute demanded by big data jobs such as \ac{ai} and genome processing, with resource requirements doubling every $3.4$ months since 2012; $50\times$ faster than Moore's Law \citep{openai2018}. This trend is showing no sign of slowing down. The fundamental relationship between neural network accuracy and scale \citep{kaplan2020} provides a strong incentive for practitioners seeking performance improvement to further increase their resource requirements. Moreover, brain-scale \ac{ai} will require at least as many parameters as the $\approx$\numprint{1000} trillion synapses present in the human brain \citep{furber2016}; several orders of magnitude more than the largest models used today. 

The compute time and memory requirements of state-of-the-art big data applications already far outstrip the capabilities of any single hardware device. For example, one of the current largest \acp{dnn}, \ac{mtnlg} \citep{megatronturingnlg2022}, contains \numprint{530} billion parameters. These parameters alone occupy $\approx$\numprint{1000} GB, exceeding the capacity of the largest A100 GPU by over an order of magnitude, and the parameter loss gradients tracked during training occupy several times more. Even if the model could be fitted onto a single device, the training time would be $\approx$\numprint{900} years\footnote{Assuming it takes \numprint{8} V100 GPUs $36$ years to train a \numprint{175} billion parameter model \citep{nvidia2022} and extrapolating.}. To address these compute time and memory demands, rather than using a single device, big data jobs must be distributed and parallelised across a cluster of machines. For example, the Selene supercomputing cluster \citep{nvidia2020selene} consists of \numprint{358400} A100 GPU tensor cores, bringing the \ac{mtnlg} training time from \numprint{900} years down to the order of days\footnote{Assuming a linear parallelisation speedup and $0$ communication overhead.}.

\begin{figure}[!tp]
    \centering
    \includegraphics[width=0.7\textwidth]{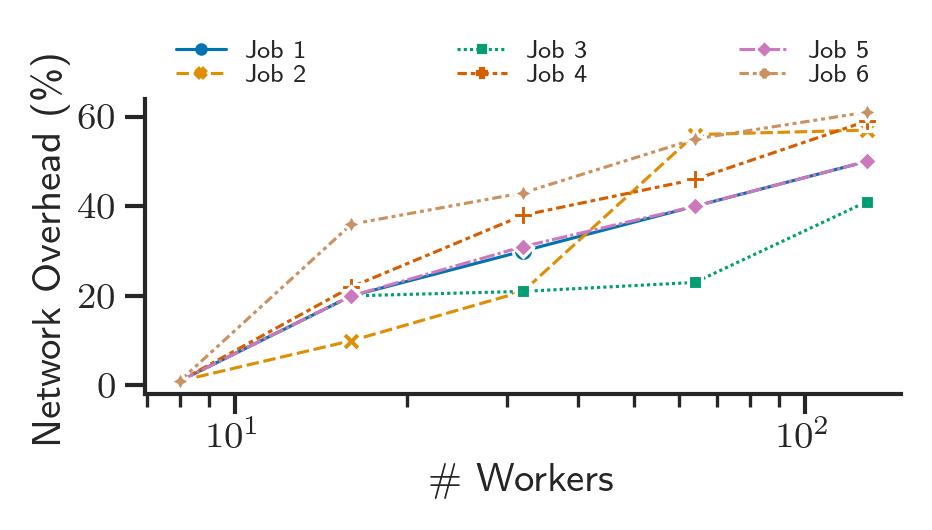}
    \caption{How the network overhead of six distributed deep learning jobs (encompassing object tracking, recommendation, natural language processing, and image recognition) increases with the number of workers used in Meta's GPU cluster \citep{wang2022topoopt}.}
    \label{fig:network_overhead_vs_number_of_workers}
\end{figure}

However, parallelising jobs across ever-more machines brings its own challenges. With any parallelisation strategy, at some point the output of each `worker' (a single device processing at least part of a job) must be collected and synchronised to get the overall result of the parallelised computation. This synchronisation requires \emph{communication} between the workers. As the number of workers used to execute a job is increased, the per-worker computation demands decrease, but the overall communication overhead between workers grows (see Figure \ref{fig:network_overhead_vs_number_of_workers}). This shifts the performance bottleneck away from the workers themselves and into the network connecting them, and brings additional challenges with managing varying traffic characteristics for different job types and parallelisation strategies \citep{wang2022topoopt, parsonson2022traffic, benjamin2021benchmarking, benjamin2022optical}.






To address the communication bottleneck in distributed computing, recent works have sought to develop optical clusters \citep{benjamin2020pulse, ballani2020sirius, khani2021sipml, wang2022topoopt, ottino2022ramp}; machines interconnected by optical switches \citep{parsonson2020optimal, gerard2020swift, gerard2021ai}. Compared to their electronic counterparts, optically switched networks offer orders of magnitude improvements in scalability, bandwidth, latency, and power consumption \citep{ballani2020sirius, zervas2018optically, vaibhawa2021monet} (see Section \ref{sec:background}). 

Optical clusters are typically operated under the \ac{ocs} paradigm due to its non-blocking circuit configurations with high capacity and scalability \citep{raja2021ultrafast}. \Ac{ocs} networks are fundamentally different from the \ac{eps} architectures used by most current clusters, resulting in entirely new communication patterns and resource demand characteristics. Consequently, new compute and network resource management schemes are needed in order to optimally allocate jobs and maximise performance.

Of the many resource management tasks which must be performed in a compute cluster, \emph{job partitioning} (how to split a job up across how many devices) is key to overall performance. More partitioning can lead to lower compute times. However, it may also increase network overhead and occupancy of cluster resources, possibly leading to future jobs being blocked upon arrival and consequently lower overall cluster throughput. Prior works such as SiP-ML \citep{khani2021sipml} have introduced simple partitioning heuristics for optical networks which have notably improved cluster performance. However, they have not been designed under the more realistic setting of dynamic and stochastic job arrivals, have not considered the state of the cluster in a `network-aware' manner when making partitioning decisions, and have been crafted to optimise for the sub-optimal objective of minimising \ac{jct}.

In this work, we first argue that simply minimising the \ac{jct} is a naive objective because it brazenly encourages more parallelisation of a job request without considering the effect this has on the ability of a cluster to service subsequent jobs. We then introduce a new more subtle formulation of the optimisation metric, the \emph{user-defined blocking rate}, which more aptly encompasses the desires of cluster users.  Next, we propose a simple modification of the quantised SiP-ML partitioner which, rather than maximally parallelising all jobs, minimally parallelises them such that they meet the user-defined maximum acceptable completion time. Then, we propose a novel network-aware partitioning strategy (see Figure \ref{fig:rl_gnn_partitioning_methodology} and Section \ref{sec:methodology}) called PAC-ML (\underline{\textbf{p}}artitioning for \underline{\textbf{a}}synchronous \underline{\textbf{c}}omputing with \underline{\textbf{m}}achine \underline{\textbf{l}}earning) which utilises \ac{rl} and a \ac{gnn} to flexibly meet the demands of the user in an arbitrary manner given the current state of the cluster network. Finally, we demonstrate our method in simulation on the recently propsed RAMP optical architecture \citep{ottino2022ramp}, achieving up to $56.2\%$ lower blocking rates than the best heuristic baseline. We show that different user-defined demand environments require different partitioning strategies for optimal results, and that a key advantage of PAC-ML is that it is able to discover performant strategies \textit{automatically} without the need for handcrafted heuristics or environment-specific tuning.

\section{Related Work}
\label{sec:related_work}


Recent years have seen a surge of interest in developing methods to distribute \ac{ml} tasks across multiple devices \citep{bennun2019demystifying, ruben2020scalable}. One approach has been to optimise the \textit{physical plane} of the distributed cluster such as its compute and network devices and architectures \citep{parsonson2020optimal, khani2021sipml, wang2022topoopt, ottino2022ramp}. In this work, we instead focus on optimising the \textit{virtual plane}, which determines how physical layer resources are allocated to execute a job. We divide the virtual plane into three sub-components: Job $(1)$ partitioning (\textit{how many} devices to use); $(2)$ placement (\textit{which} devices to use); and $(3)$ scheduling (\textit{in which order} to use the devices). Many prior virtual plane works have considered $(2)$ and $(3)$ (\textit{how} to distribute), whereas we focus on $(1)$ (\textit{how much} to distribute). However, in this section we comment on recent progress across all these fields, since we leverage this progress throughout our work.

\secInLine{ML for discrete optimisation}
Many \ac{co} problems turn out to be NP-hard, rendering exhaustive search techniques intractable for practical application \citep{bengio2021machine}. Consequently, practitioners rely on either approximate algorithms, which give restricted performance guarantees and poor scalability \citep{williamson2011design}, or heuristics, which have limited solution efficacy \citep{Halim2019CombinatorialOC}. Since the first application of neural networks to \ac{co} by \cite{hopfield1985neural}, the last decade has seen a resurgence in ML-for-CO \citep{bello2017neural, dai2017hanjun, barrett2019exploratory, gasse2019exact, barrett2022learning, parsonson2022retro}. The advantages of ML-for-CO over approximation algorithms and heuristics include handling complex problems at scale, learning either without external input and achieving super-human performance or imitating strong but computationally expensive solvers, and (after training) leveraging the fast inference time of a \ac{dnn} forward pass to rapidly generate solutions. Since almost all cluster resource management tasks can be reduced to canonical \ac{co} problems \citep{bengio2021machine}, many state-of-the-art resource management methods utilise recent advances in \ac{ml}-for-\ac{co}.

\secInLine{Job placement}
\cite{mirhoseini2017device} were the first to apply \ac{ml} to the task of deciding which operations in a computation graph to place on which devices in a cluster. They used a sequence-to-sequence model consisting of an LSTM \ac{dnn} with an attention mechanism trained with the simple REINFORCE policy gradient \ac{rl} algorithm \citep{williams1992simple} such that the \ac{jct} of a deep learning job was minimised, outperforming handcrafted heuristics when training the Inception-V3 computer vision and LSTM natural language processing models. \cite{gao2018spotlight} furthered this work by replacing REINFORCE with the more advanced \ac{ppo} \ac{rl} algorithm \citep{schulman2017ppo} with lower variance and reduced training hardware demands. They demonstrated their method beating \cite{mirhoseini2017device} on the CIFAR-10 image recognition benchmark in terms of \ac{jct}. \cite{mirhoseini2018a} proposed a novel hierarchical model which decomposed the job placement task into a joint group-and-place problem, reducing the \ac{jct} of Inception-V3, ResNet, LSTM, and NMT models by up to $60\%$ relative to the state-of-the-art. 

All works up to this point used \ac{dnn} architectures restricted to Euclidean-structured input data. Consequently, in order to handle non-Euclidean graph-structured data such as computation graphs and cluster networks, they had to be re-trained each time a new graph structure was considered. \cite{addanki2019placeto} were the first to instead leverage a \ac{gnn}, as well as the grouping scheme of \cite{mirhoseini2018a}, to learn to generalise across different job types with varying computation graph structures, demonstrating device placement schemes which were on par with or better than prior approaches on Inception-V4, NASNet, and NMT after $6.1\times$ fewer training steps. \cite{khadka2021optimizing} furthered the use of \acp{gnn} for job placement by combining \acp{gnn}, \ac{rl}, and population-based evolutionary search with the hierarchical group-and-place scheme of \cite{mirhoseini2018a}. Concretely, they replaced the manually-designed operation grouping heuristic with a learned policy capable of superior scaling and \ac{jct} performance.


\secInLine{Job scheduling}
\cite{bao2018online} addressed the job scheduling problem (the order in which to execute operations placed across a set of devices) using a primal-dual framework for online job scheduling. They represented the problem as an \ac{ilp} which their proposed algorithm could solve in polynomial time in an online fashion such that the cluster resources were maximally utilised and the \ac{jct} minimised. \cite{li2021scheduling} proposed a placement-aware scheme which leveraged the pre-determined device placement allocation to decide on a job schedule which could reduce the average \ac{jct} by up to $25\%$ relative to other scheduling methods. \cite{paliwal2020reinforced} went further by utilising an \ac{rl}-trained \ac{gnn} and a genetic algorithm to jointly optimise both job placement and scheduling, demonstrating both lower \ac{jct} and peak memory usage than other strategies when distributing TensorFlow computation graphs across a cluster.

\secInLine{Job partitioning}
To the best of our knowledge, \cite{khani2021sipml} are the only ones to have explicitly considered the question of \textit{how much} to distribute a computation graph in the context of an optical network. Like other works, they assumed that a maximum parallelisation strategy (i.e. partition the job across as many workers as possible) is a desirable objective, and then focused on how best to design the physical layer such that the \ac{jct} could be minimised.

All works discussed in this section have assumed that the \ac{jct} is the key objective to minimise. Consequently, where the question of partitioning is considered, prior works have assumed that more parallelisation is desirable. However, we posit that user-critical metrics such as throughput and blocking rate are compromised by prioritising optimisation of the \ac{jct} in a cluster setting with dynamic job arrivals. To address this shortcoming, we propose a new \ac{ml}-based resource management scheme which explicitly addresses the partitioning question. Concretely, our work leverages the emergent trend from these other virtual plane fields, namely utilising an \ac{rl}-trained \ac{gnn}, to decide how much to partition different jobs in a dynamic setting with arbitrary user-defined completion time requirements.

\section{Background}
\label{sec:background}

\begin{figure}[!tp]
    \centering
    \includegraphics[width=0.6\textwidth]{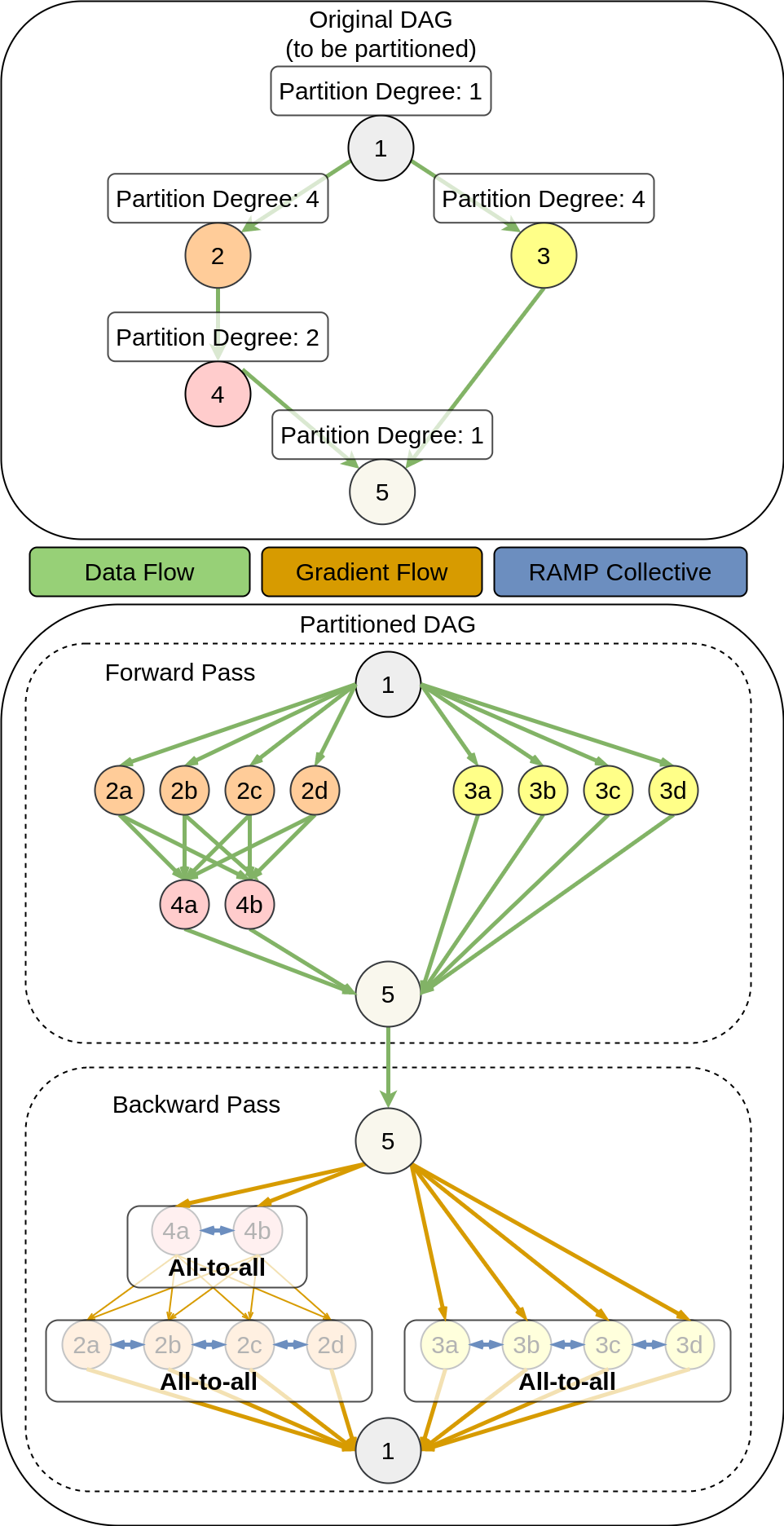}
    \caption{Diagram showing a \ac{dnn} job DAG being partitioned. Top: A forward pass DAG where each node has an associated partition degree (how many times it will be divided when partitioned). Bottom: A partitioned DAG with forward and backward passes handled consecutively. Green edges in the graph represent data flow (i.e. output to input) between consecutive operations in the forward pass. Orange edges represent gradient exchanges processed in the backward pass (backpropagation). Blue edges represent full connectivity collective operations to synchronise weight updates across partitioned components of an operation. Note that, for brevity, the top unpartitioned DAG only shows the forward pass (since, before partitioning, the graph structure is identical to the backward pass), whereas the bottom partitioned DAG shows both the forward and backwards passes (since, after partitioning, the graph structures are different).}
    \label{fig:dag_partition}
\end{figure}

\subsection{Parallelisation}

\secInLine{Types of parallelism}
Parallelisation is the process of distributing a computational job across multiple devices. This is done in order to reduce the time and/or physical memory needed to complete the job. There are three main types of deep learning parallelism; \textit{data parallelism}, \textit{model parallelism}, and \textit{hybrid parallelism} (see Appendix \ref{sec:extended_background_parallelisation} for extended background information on these methods). Although today the most common method for \ac{dnn} training parallelisation is data parallelism for its simplicity and limited network overhead, we focus on the less common but more desirable model parallelism paradigm for its strong scaling capabilities \citep{khani2021sipml}. Our proposed partitioning methods are applicable to hybrid and pipeline parallelism, but these require additional simulation complexity and are therefore beyond the scope of this manuscript.

\secInLine{Computational jobs}
A computational job is a \ac{dag} whose nodes are \textit{operations} and edges are \textit{dependencies}. Operations are computational tasks (e.g. some mathematical reduction, a database query, etc.). Dependencies are either \textit{control dependencies}, where the child operation can only begin once the parent operation has been completed, or \textit{data dependencies}, where at least one tensor is output from the parent operation and required as input to the child operation. In the context of \acp{dnn}, a job \ac{dag} is a sequence of forward pass, backward pass, and parameter update operations which need to be performed on data exchanged between operations. Whether or not this data passes through a communication network is determined by how the operations are partitioned, placed across a cluster of workers, and parallelised.  

\secInLine{Job partitioning}
Job partitioning refers to the process of splitting the operations of a job \ac{dag} into $u$ (the \textit{partition degree}) smaller sub-operations which can in turn be placed across $u$ workers, thus reducing their run time and memory requirements. Partitioning is used in the model, hybrid, and pipeline parallelisim paradigms. More partitioning can decrease compute time and memory requirements, but requires more inter-worker communication, complex intra-worker operation scheduling, and greater resource utilisation, therefore potentially increasing overall completion time, cluster complexity, and subsequent job blocking rates. Figure \ref{fig:dag_partition} visualises how an initial DAG for some arbitrary neural network architecture, where each operation has a partitioning degree, can be re-represented in terms of its partitioned form. Both forward and backward passes are explicitly represented since inter-operation information dependencies (i.e. the edges in the graph) are not the same in each pass.



\subsection{Optical Networking}
Most current cluster networks use optic fibre communication links, but the switch devices which interconnect the network are usually electronic.

\secInLine{Limitations of electronic networking}
Electronic networks have poor scalability, bandwidth, latency, and power consumption. Concretely, since the per-port bandwidth is limited and the power consumption required to cool active electronic devices is expensive, the bisection bandwidth achievable in an electronic network is restricted, thus hampering scalability. Consequently, although the compute power of DCN server nodes, as measured by FLOP/s, has increased by a factor of \num{65} over the last \num{18} years, the bandwidth of the DCN network facilitating communication between these nodes has only increased by a factor of \num{4.8}, resulting in an 8-factor decrease in bytes communicated per FLOP \citep{Bergman2018empowering}. This has created a performance bottleneck not in the server nodes themselves, but rather in the network connecting them. This issue is especially compounded when striving for strong scaling via model parallelism with distributed computing, and with the trend towards larger models with ever more parameters as described in Section \ref{sec:introduction}.

\secInLine{Optical circuit switched networks}
Cluster networks with optical switches have the potential to offer significant improvements in performance (due to larger bandwidth and lower switching latency) and energy efficiency (due to the lack of optical-electronic-optical conversion overhead), as well as the capability to scale to next-generation large-scale distributed compute jobs with exascale bandwidth and compute \citep{ottino2022ramp}. \Ac{ocs} networks in particular offer a promising avenue with which to realise commercial optical networks due to their non-blocking circuit configurations with high capacity and scalability and low deterministic switching latency. In contrast to optical packet switched networks, \ac{ocs} networks are simpler to implement and they eliminate the need for in-switch buffering or queuing and addressing.

\begin{figure}[!tp]
    \centering
    \includegraphics[width=0.7\textwidth]{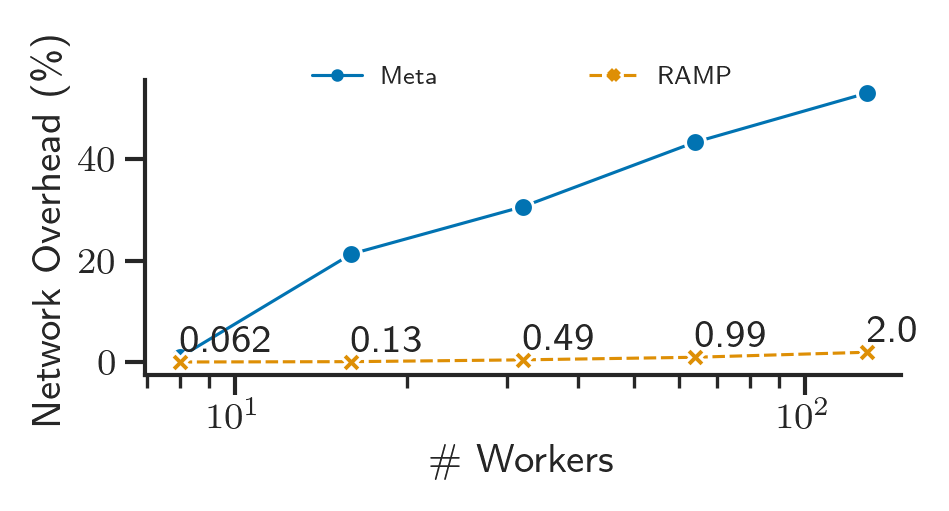}
    \caption{The mean network overhead of the \num{6} distributed deep learning jobs reported by \citep{wang2022topoopt} in Meta's GPU cluster compared to that of RAMP as reported by \cite{ottino2022ramp} on the \num{5} jobs considered in our work. Note that this is an approximate comparison, and that the important takeaway is that RAMP retains low network overheads as jobs become increasingly distributed.}
    \label{fig:ramp_vs_meta_network_overhead}
\end{figure}

\secInLine{RAMP}
RAMP is a state-of-the-art \ac{ocs} architecture designed specifically for cloud data centres and distributed deep learning systems \citep{ottino2022ramp}. RAMP networks are parameterised by $N_{C}$ communication groups, $N_{R}$ racks per communication group, and $N_{S}$ servers per rack, resulting in a $N_{W}=N_{C} \times N_{R} \times N_{S}$ worker cluster with a colloquially termed `RAMP shape' defined by tuple $\langle N_{C}, N_{R}, N_{S} \rangle$. At its core, RAMP proposes a novel set of \acp{mpi} for performing the synchronisation steps (AllReduce, AllGather, etc.) required by distributed \ac{dnn} training jobs. These will be referred to as \textit{collective operations}. These \acp{mpi} are designed to take full advantage of the high bandwidth provided by optical network architectures. Consequently, as shown in Figure \ref{fig:ramp_vs_meta_network_overhead}, the network overhead of RAMP remains remarkably low as the number of workers used to execute a job increase (see Section \ref{sec:experimental_setup} for experimental details). The RAMP authors showed that this low network overhead enables unprecedented scalability with up to \num{65536} worker nodes capable of training $O(\text{trillion})$ parameter \ac{dnn} models.

\secInLine{RAMP placement rules}
As detailed in \cite{ottino2022ramp}, a group of workers in a RAMP shape can only undergo collective operations if they are selected with respect to certain rules, loosely termed here `symmetry' rules. For shape $\langle N_{C}, N_{R}, N_{S} \rangle$, these rules are as follows: $(1)$ $N_{S}$ workers per rack spread over $N_{R}$ racks requires that the set of workers on each rack span $N_{R}$ distinct communication groups. These $N_{R}$ distinct communication groups do not have to be the same set across racks. $(2)$ $N_{S}$ workers on $N_{R} = 1$ rack must span $N_{S}$ communication groups. $(3)$ $N_{S}$ workers spread over $N_{R}$ racks ($N_{S} = 1$ worker per rack) must span $N_{S}$ distinct communication groups.

In our simulations, we use a simple first-fit operation placement heuristic which conforms to these rules (refer to Appendix \ref{sec:first_fit_operation_placement_in_ramp} for further details).

\subsection{Reinforcement Learning}

\begin{figure*}[!tp]
    \centering
    \includegraphics[width=0.98\textwidth]{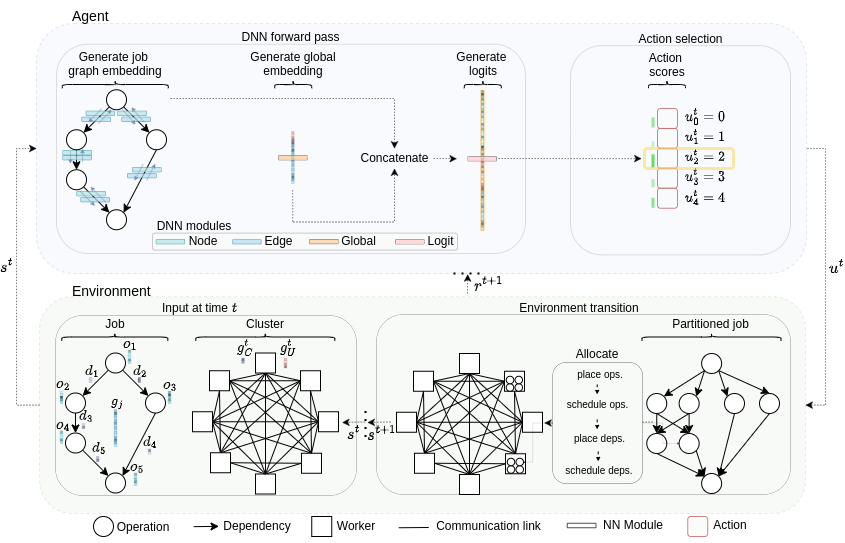}
    \caption{An overview of our PAC-ML approach transitioning from step $t \rightarrow t + 1$. At each time step $t$ when there is a new job to be placed on the cluster, we: (i) Use a \ac{gnn} to generate an embedded representation of the node and edge features in the job's computation graph, and a standard feedforward \ac{dnn} to do the same for the global job and cluster features; (ii) concatenate the outputs of (i) and use another feedforward \ac{dnn} to generate a logit for each action $u^{t} \in U^{t}$; (iii) pass the chosen action $u^{t}$ to the environment and partition the job accordingly; (iv) apply any internal environment allocation heuristics (operation and dependency placement and scheduling, etc.) to attempt to host the job on the cluster; (v) if accepted onto the cluster, perform a lookahead to evaluate the job's completion time; (vi) fast-forward the environment's wall clock time $t_{\text{wc}}$ to when the next job arrives, and return the corresponding reward $r^{t+1}$ and updated state $s^{t+1}$.}
    \label{fig:rl_gnn_partitioning_methodology}
\end{figure*}

\Ac{rl} is the study of optimal decision making in natural and artificial systems \citep{sutton1998reinforcement}. In the general \ac{rl} setting shown in Figure \ref{fig:rl_gnn_partitioning_methodology}, an \textit{agent} interacts with an \textit{environment} at each sequential time step $t$. The environment can be described by tuple $\langle T, R \rangle$, where $T$ is a state transition probability matrix defining the transition probabilities from all states $s$ to all successor states $s'$ taking action $u$ where $T^{u}_{ss'} = \mathbb{P}(S^{t+1}=s'|S^{t}=s, U^{t}=u)$, and $R$ is a scalar reward function giving the expected immediate (next state) reward given current state $s$ and chosen action $u$ where $R^{u}_{s} = \mathbb{E}(R^{t+1} | S^{t} = s, U^{t} = u)$.

\secInLine{Markov decision process}
The environment is usually assumed to have the \textit{Markov property} whereby $\mathbb{P}(s^{t+1} | s^{t}) = \mathbb{P}(s^{t+1} | h^{t})$; that is to say that the probability of the next state being $s^{t+1}$ given the current state $s^{t}$ is the same as the equivalent probability given all previous states in history $h^{t} = \{s^{1}, ..., s^{t}\}$. As such, this RL setting is usually assumed to be a \ac{mdp} described by tuple $\langle S, U, T, R, \gamma \rangle$ where $S$ is a finite set of possible environment states, $U$ is either a discrete (finite) or continuous (infinite) set of possible actions, and $\gamma \in [0, 1]$ is a discount factor specifying the factor by which to multiply future expected rewards to discount their present value. Since Markov states are stochastic, future rewards are never fully certain and are therefore expressed as an \textit{expectation}. 

\secInLine{Agent goal}
The agent's goal is to learn to maximise its expected total discounted future reward, termed the `value' or `return' $G^{t}=\sum_{k=0}^{\infty} \gamma^k R^{t+k+1}$, over the course of an \textit{episode} (a sequence of decision steps which may or may not terminate at some point). To do so, the agent can use \textit{model-free} \ac{rl} to avoid explicitly modelling the environment by only using its \textit{policy function} and/or its \textit{value function} to make decisions. The policy function $\pi$ maps an observed state $s^{t}$ to a corresponding action $u^{t}$ such that some estimated score objective is maximised. The value function estimates the expected return $G_{t}$ from being in state $s^{t}$ and following policy $\pi$ (the \textit{state value function} $v$) or from being in state $s^{t}$, taking action $u^{t}$, and following policy $\pi$ (the \textit{action value function} $q$). Crucially, value and policy functions can be approximated and learned with \acp{dnn}, enabling \ac{rl} to be scaled to large problem instances (see Appendix \ref{sec:extended_background_neural_networks} for extended background information on \acp{dnn}).

\secInLine{Advantages of RL}
Using traditional \ac{rl} has several advantages over heuristics and other \ac{ml} paradigms such as supervised learning. First, no external data from human-designed or computationally expensive heuristics is required, enabling an agent to learn super-human policies without potentially sub-optimal initial biases towards a certain strategy or a costly expert example collection-and-labelling phase \citep{SilverHuangEtAl16nature}. Second, a \ac{dnn} with a finite number of layers and neurons will have its expressivity constrained \citep{dong2020on}, restricting the complexity of the set of functions it is capable of approximating. Because the objective of an \ac{rl} agent is to maximise its expected future return which, under the assumption that a suitable reward function has been crafted, is equivalent to maximising performance on a given task, \ac{rl} agents are able to maximise task performance given \ac{dnn} expressivity constraints. Third, since \ac{rl} agents maximise \textit{future} return, they are capable of learning sophisticated non-myopic policies which sacrifice short-term reward in exchange for higher long-term return \citep{sutton1998reinforcement}.



\section{User-Defined Blocking Rate}
\label{sec:user_defined_blocking_rate}

To motivate our work, we first explore the key metrics to consider when evaluating a job partitioning strategy with the help of an experiment on $32$ GPU workers, and then introduce a new formulation of the \textit{user-defined blocking rate}. All experimental details are given in Section \ref{sec:experimental_setup}.

\begin{figure}[!tp]
    \centering
    \includegraphics[width=0.7\textwidth]{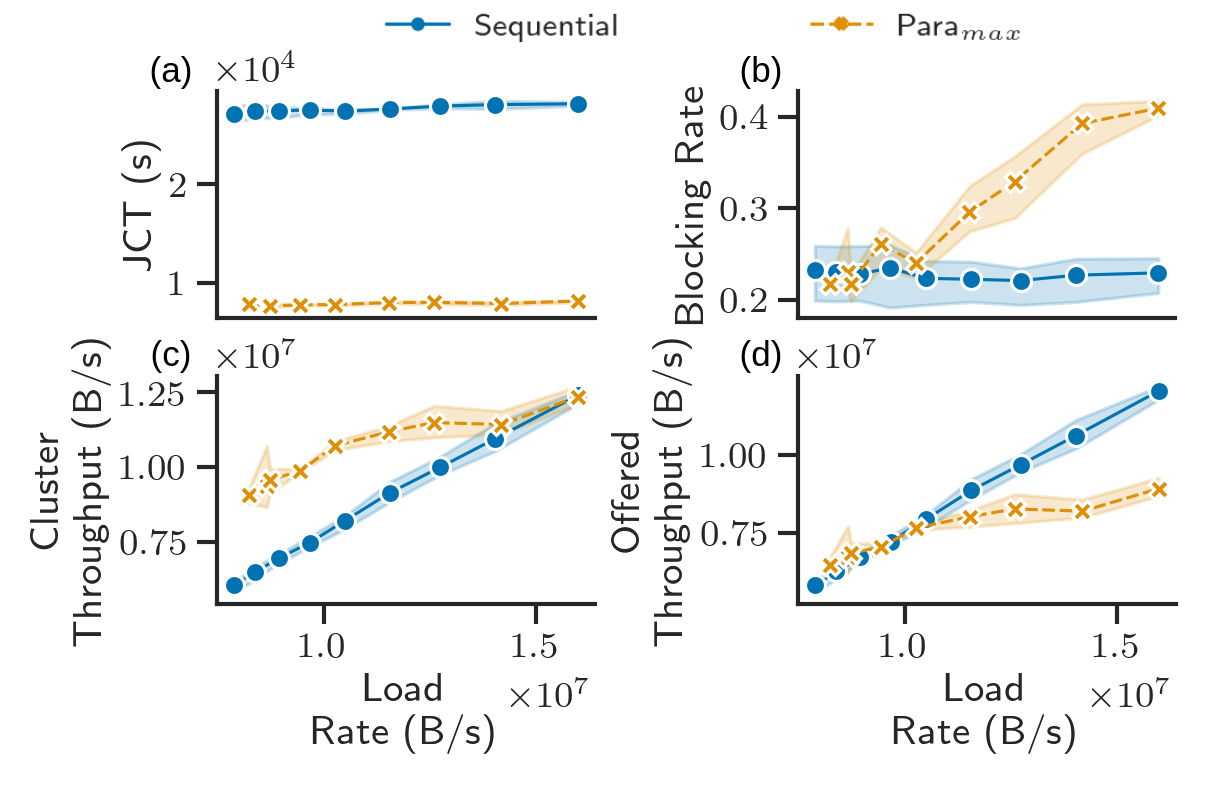}
    \caption{(a-b) Demonstration of how more partitioning can lead to a lower \ac{jct} than no partitioning (i.e. sequentially running the job on a single device), but this may be at the cost of a higher blocking rate since more cluster resources are occupied when subsequent jobs arrive. (c-d) Demonstration of how optimising for the cluster throughput leads to an unfair bias towards more partitioning, because more parallelism creates more work for the cluster and therefore artificially increases cluster throughput even though, from the perspective of the user, the original offered throughput may be lower.}
    \label{fig:inadequate_metrics_demo}
\end{figure}



\secInLine{The inadequacy of optimising the job completion time}
As discussed in Section \ref{sec:related_work}, most prior works researching management schemes for distributed computing aim to minimise \ac{jct}; the time taken to complete a given job. If a job $j$ begins running at wall clock time $t_{wc, j}^{\text{start}}$ and is completed at time $t_{wc, j}^{\text{end}}$, researchers usually record the completion time as $\text{JCT}_{j} = t_{wc, j}^{\text{end}} - t_{wc, j}^{\text{start}}$. Consequently, most systems maximise the degree to which they parallelise jobs in order to minimise \ac{jct}. While it is true that end users undoubtedly want this \ac{jct} metric to be minimised, it fails to quantify when a job was \textit{blocked}, which occurs when no cluster resources were available to service it. While more parallelism will often lead to a lower \ac{jct} for a given job, it will also use up more of the cluster's compute and network resources, potentially blocking future job arrivals (see Figure \ref{fig:inadequate_metrics_demo}). Therefore in practice, end-users wish to minimise both the \ac{jct} and the overall \textit{blocking rate} (the fraction of jobs blocked over a given time period). While maximum parallelisation will lead to a minimised \ac{jct}, we posit that a balance between these two extreme parallelisation strategies can more aptly optimise for both the \ac{jct} and blocking rate.



\secInLine{Alternative optimisation objectives}
One metric which encapsulates both the \ac{jct} and blocking rate is \textit{throughput}; the information processed per unit time. There are two issues with using throughput as an optimisation objective. (1) Operators must be careful how they measure the throughput to be optimised. If they measure the \textit{cluster throughput} (the total \textit{cluster information} processed per unit time), they will be biased towards more parallelisation, because when a job is partitioned and parallelised, the edge dependencies coming in to and out of the partitioned operation node(s) must be replicated (see Figure \ref{fig:dag_partition}). This artificially creates more information for the cluster to process even though, from the end users' perspective, the total information processed of their original demand is the same. Therefore, the \textit{offered throughput} (the total original demand information (i.e. before partitioning was applied) processed per unit time) is a more suitable throughput metric to optimise. Figure~\ref{fig:inadequate_metrics_demo} shows an example of how a `maximum partitioning' strategy, such as that used by SiP-ML \citep{khani2021sipml}, can have superior cluster throughput when compared to a `no partitioning' strategy (sequentially running the job on a single device) despite having lower offered throughput. However, offered throughput is still an inadequate optimisation metric, because (2) in practice, different jobs being serviced by the cluster originating from different client users have different priorities and job completion time requirements. For example, two identical machine learning training jobs might be submitted to the cluster, but one from a user who intends to deploy the model commercially and requires it to be completed overnight, and the other from a user who is employing the model for research and has less stringent completion time requirements. Ideally, operators would direct their clusters to meet flexible user-defined per-job completion time requirements.


\secInLine{The user-defined blocking rate}
To enable users to dynamically determine the completion time on a per-job basis whilst also maximising the number of job demands satisfied, we introduce a new formulation of the \textit{user-defined blocking rate} objective for the partitioning algorithm to optimise. Given a job which, if executed sequentially on one device, would be completed in $\text{JCT}_{j}^{\text{seq}}$, we define the \textit{maximum acceptable \ac{jct}} as $\text{JCT}_{j}^{\text{acc}} = \beta \cdot \text{JCT}_{j}^{\text{seq}}$, where $\{ \beta \in \mathbb{R}: 0 < \beta \leq 1 \}$. Here, $\beta$ is a parameter chosen by the user which determines how quickly the job must be completed. If $\text{JCT}_{j} > \beta \cdot \text{JCT}_{j}^{\text{seq}}$, then the cluster will have failed to complete the job within the required time and the job will be recorded as having been blocked. The user-defined blocking rate is therefore the fraction of jobs which failed to meet the $\text{JCT}_{j} \leq \beta \cdot \text{JCT}_{j}^{\text{seq}}$ requirement over a given period of time. Note that rather than brazenly optimising for either the \ac{jct} or the blocking rate, the user-defined blocking rate enables the cluster operator to instead dynamically specify their desired completion time on a per-job basis, and the performance of the cluster is evaluated according to how well it was able to meet the requirements of the user. Furthermore, the $\beta$ parameter corresponds to the speed-up factor being requested by the user and, since $\{ \beta \in \mathbb{R}: 0 < \beta \leq 1 \}$, can be given directly as input to a \ac{dnn}.

\section{PAC-ML Partitioning Methodology}
\label{sec:methodology}

\Ac{rl} agents can learn general policies without the need for human guidance. An \ac{rl} job partitioner therefore has the potential to take an arbitrary maximum acceptable \ac{jct} provided by the user and \textit{automatically} decide how much to distribute the job such that, over a period of time, the number of jobs which meet the \ac{jct} requirements specified by the user is maximised. Such an agent would therefore be able to minimise the blocking rate whilst also accounting for the flexible and dynamic \ac{jct} specifications of the user. Following this logic, we now describe our PAC-ML (\underline{\textbf{p}}artitioning for \underline{\textbf{a}}synchronous \underline{\textbf{c}}omputing with \underline{\textbf{m}}achine \underline{\textbf{l}}earning) approach for learning to partition computation jobs with \ac{rl} and a \ac{gnn}.

\subsection{Markov Decision Process Formulation}

Since allocating cluster resources for jobs arriving dynamically in time is a sequential decision making process, formulating problems such as job partitioning as an \ac{mdp} is a natural approach and facilitates the application of many traditional and state-of-the-art \ac{rl} algorithms \citep{mao2016resource, ravichandra2019placeto, paliwal2020reinforced}.

\secInLine{States}
A new job $j$ arriving at time step $t$ is comprised of a \ac{dag} $G(O, D, g_{j})$ with node operations $O$, edge dependencies $D$, and any other job statistics which might be recorded $g_{j}$. Similarly, the state of the cluster at time $t$ is made up of the number of workers available, the jobs currently running on the cluster, and so on. To compress the state of the cluster and the job requesting to be placed into a representation suitable as input for a neural network at time step $t$, we encode this information into five feature vectors: 

\begin{enumerate}
    \item \textbf{Per-operation features} $o_{i} \forall i \in \{1, ..., |O|\}$ ($5$ features): (i) The compute cost (run time in seconds on an A100 GPU); (ii) a binary variable indicating whether the operation has the greatest compute cost in the job; (iii) the memory cost (byte occupancy); (iv) a binary variable indicating whether the operation has the greatest memory cost in the job; and (v) the node depth with respect to the source node. The compute and memory costs are normalised by the highest compute and memory cost operations in the job, and the node depth is normalised by the depth of the deepest node.
    
    \item \textbf{Per-dependency features} $d_{i} \forall i \in \{1, ..., |D|\}$ ($2$ features): (i) The size (in bytes) of the edge dependency normalised by the largest dependency in the job; and (ii) a binary indicator of whether the dependency is the largest in the job.
    
    \item \textbf{Global job features} $g_{j}$ ($15$ features): (i) The number of operations; (ii) the number of dependencies; (iii) the sequential job completion time; (iv) the maximum acceptable job completion time; the maximum acceptable job completion time fraction $\beta$ both (v) raw and (vi) normalised; (vii) the total memory cost of all of the operations; (viii) the total size of all of the dependencies; (ix) the number of training steps which need to be performed; the (x) mean and (xi) median of the operation compute costs; the (xii) mean and (xiii) median of the operation memory costs; and (xiv) the mean and (xv) median of the dependency sizes. Each feature is normalised by the highest respective value of the feature across all job types.
    
    \item \textbf{Global cluster features} $g^{t}_{C}$ ($2$ features): (i) The number of occupied workers; and (ii) the number of jobs running. Both features are normalised by the total number of workers in the cluster $N_{W}$.
    
    \item \textbf{Global action features} $g^{t}_{U}$ ($\frac{N_{W}}{2}$ features): A binary vector indicating the validity of each possible partitioning decision given the state of the cluster and the RAMP rules defined by \citep{ottino2022ramp}.
    
\end{enumerate}


\secInLine{Actions}
Given the state $s^{t}$ encapsulating both the job requesting to be placed and the current state of the cluster, the partitioning agent uses a policy $\pi(s^{t})$ to select a number of times $u^{t}$ up to which to partition each operation in the job's computation graph (using a similar minimum operation run time quantum discretisation scheme to \cite{khani2021sipml}), where $u^{t}_{i} \forall i \in \{ 0, 1, ..., \frac{N_{W}}{2} \}$ (i.e. there are $ \big( \frac{N_{W}}{2} + 1 \big)$ possible discrete actions). Note that $u^{t}=0$ enables the agent to reject a job without placing it, $u^{t}=1$ places the job onto one worker and runs it sequentially, and $1 < u^{t} \leq \frac{N_{W}}{2}$ attempts to distribute the job's operations across up to $u^{t}$ workers. In our setting and given the RAMP rules of \cite{ottino2022ramp}, an invalid partitioning action is one which is at least one of: (i) An odd number (except $u^{t}=1$), or either (ii) greater than the number of workers available or (iii) has no valid RAMP placement shape given the current state of the cluster (see Section \ref{sec:background}).

\secInLine{Rewards}
As a consequence of the RAMP rules defined by \cite{ottino2022ramp}, which require that the worker and network resources allocated to a given job are reserved exclusively for that job for the duration of its run time, we are able to perform a deterministic lookahead to evaluate what the overall completion time, $\text{JCT}_{j}$, of the job will be as soon as it is placed. Subsequently, when a job $j$ arrives at time step $t$, we can immediately determine whether or not the cluster met the $\text{JCT}^{\text{acc}}_{j}$ specified by the user. This enables the use of a simple per-step $+1$/$-1$ reward scheme,

\begin{equation}
    r^{t+1} = 
    \begin{cases}
        1,& \text{if } \text{JCT}_{j} \leq \beta \cdot \text{JCT}_{j}^{\text{seq}} \\
        -1,              & \text{otherwise}
    \end{cases},
\end{equation}

which when aggregated and maximised over the course of an episode corresponds to maximally meeting the specified per-job completion time requirements and therefore minimising the user-defined blocking rate.

\secInLine{Transitions}
In our hybrid time- and event-drive simulation, when the agent makes a partitioning decision at time step $t$, the environment transitions to the next step $t+1$ by fast-forwarding its internal simulated wall clock time, $t_{\text{wc}}$, to when the next job arrives and requests to be placed, updating the states of any running and completed jobs and their corresponding compute and network resources as necessary. The episode terminates when $t_{\text{wc}} = T_{\text{wc}}^{\text{max}}$.

\subsection{PAC-ML Learning Setup}

\secInLine{Reinforcement learning algorithm}
To find a policy which maximises the expected return when partitioning jobs, we used the state-of-the-art Ape-X DQN \citep{horgan2018distributed} \ac{rl} algorithm; a distributed and highly scalable value-based method (see Appendix \ref{sec:reinforcement_learning_algorithm} for algorithm details and hyperparameters).

\secInLine{Neural network architecture}
To make the learning of value and policy functions tractable in large state-action spaces, we approximated them with a custom-built message passing \ac{gnn} implemented using the open-source PyTorch \citep{paszke2019pytorch} and DGL \citep{wang2019dgl} libraries. Refer to Appendix \ref{sec:neural_network_architecture} for further architectural details.

\section{Experimental Setup}
\label{sec:experimental_setup}

All code for reproducing the experiments and links to the generated data sets are provided at \url{https://github.com/cwfparsonson/ddls}.

\begin{figure}[!tp]
    \centering
    \includegraphics[width=0.7\textwidth]{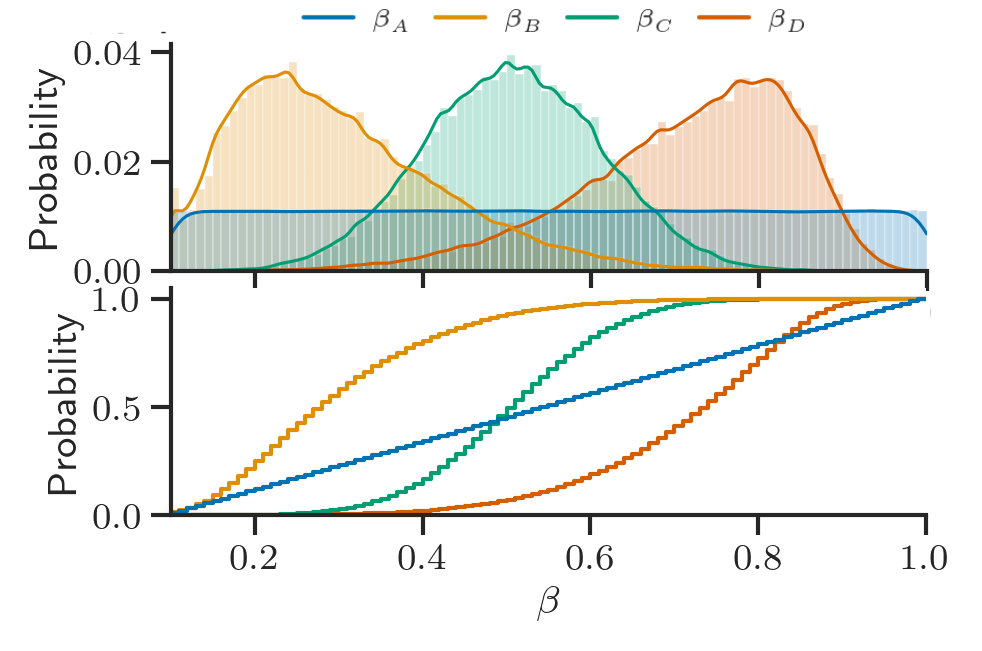}
    \caption{The four $\beta$ distributions used in our experiments in order to measure the capability of each partitioner to cater to different user-defined maximum acceptable completion time requirement settings. In each $\beta_{X}$ experiment setting, each new job generated was assigned a $\beta$ value sampled from $\beta_{X}$ in order to get the maximum acceptable job completion time, $\beta \cdot \text{JCT}^{\text{seq}}$ (see Section \ref{sec:user_defined_blocking_rate}).}
    \label{fig:training_evaluation_beta_distributions}
\end{figure}

\secInLine{Simulation environment}
We built an open-source Gym environment \citep{brockman2016openai} to simulate the RAMP \ac{ocs} system of \cite{ottino2022ramp} in an \ac{rl}-compatible manner. We used a hybrid time- and event- driven simulation approach where we kept track of the internal simulation wall clock time $t_{wc}$, enabling the measurement of time-based metrics, but only took a partitioning decision when needed (i.e. when a new job demand arrived at the cluster), aiding efficiency since no discrete steps were needlessly simulated. All our experiments used similar cluster parameters to \cite{ottino2022ramp}. We used 
$N_{W}=32$ $(N_{C}=4, N_{R}=4, N_{S}=2)$ 
A100 GPUs with $80$ GB memory capacity, $2$ THz memory frequency, and a peak computational power of $130$ Tflop/s. We assumed 
an intra-GPU propagation latency of \num{50} ns, a negligible \ac{ocs} circuit reconfiguration latency of \num{1} ns, a worker input-output latency of $100$ ns, and a total worker communication capacity of \num{1.6} TB/s (resulting in a per-transceiver bandwidth of $\frac{1.6\times10^{12}}{N_{C}}$ B/s). All experiments were run up to a simulated wall clock time of $T_{\text{wc}}^{\text{max}} = 10^{6}$ s (around $12$ days) of continuous cluster operation with dynamic job arrivals and were repeated across $3$ random seeds, with the subsequent min-max confidence intervals for each measurement metric reported. More details of the simulation environment are provided in Appendix \ref{sec:additional_simulation_environment_details}.

\secInLine{Compute jobs} 
We used the computation graph time and memory profiles of five real deep learning job types open-accessed with Microsoft's PipeDream research \citep{narayanan2019pipedream, narayanan2021memory-efficient} (see Appendix \ref{sec:job_computation_graph_data_sets} for details). These jobs encompassed image classification (AlexNet \citep{krizhevsky2012imagenet}, ResNet-18 \citep{he2016deep}, SqueezeNet-10 \citep{iandola2016squeezenet}, and VGG-16 \citep{simonyan2014very}) and natural language processing (GNMT \citep{wu2016google}) tasks, thereby testing the generality of the approaches we considered. All jobs arrived to the cluster dynamically and stochastically throughout the simulation period, with the inter-arrival time fixed at \num{1000} s to control the load rate. Each job was ran for $N_{iter}=50$ training iterations, where one training iteration consists of one forward and backward pass through the neural network.

\secInLine{Partitioning} 
When partitioning the operations in a job's computation graph, we allowed the partitioning agents to split each operation up to $\frac{N_{W}}{2}$ times (the environment's `maximum partitioning degree'). We followed \cite{khani2021sipml} by $(1)$ assuming a linear dependency between the total number of operation splits and each split's compute time; and $(2)$ choosing a minimum quantum of computation time, $\tau$, and splitting operations up to a number of times which would result in sub-operations with a compute time no smaller than $\tau$ in order to maximise GPU utilisation. We set $\tau=10$ ms. As such, a given partitioning action $u^{t}$ set the maxmimum partitioning degree of the job, but individual operations within the job could be split fewer times depending on their initial compute time and $\tau$. Note that although this restricts each operation to be distributed across a maximum of $u^{t}$ servers, the total number of workers used by all operations in the job can still be greater than $u^{t}$ depending on the operation placement heuristic's choices.

\secInLine{Maximum acceptable job completion times}
In our setting, a partitioner would ideally be able to take an arbitrary job with an arbitrary maximum acceptable job completion time, $\beta \cdot \text{JCT}^{\text{seq}}$, and partition the job such that the completion time requirement is satisfied for as many dynamically arriving jobs as possible (thereby minimising the user-defined blocking rate; see Section \ref{sec:user_defined_blocking_rate}). To test each partitioner's ability to do this, we ran experiments using four $\beta$ distributions ($\beta_{A}, \beta_{B}, \beta_{C}, \text{ and } \beta_{D}$; see Figure \ref{fig:training_evaluation_beta_distributions}). For each $\beta_{X}$ experiment, when one of the five possible jobs was randomly generated to arrive at the cluster, a $\beta$ value, discretised to two decimal places, was randomly sampled from the experiment's $\beta_{X}$ distribution and assigned to the job. By sampling a broad range of $\beta$ values from a selection of $\beta_{X}$ distributions, we ensured that we could analyse the performance of each partitioning agent under different completion time requirement settings and subsequently measure the capability of each method to cater for different user-defined requirements.

\begin{table}[!htp]
    \centering
    \begin{tabular}{ l c c c | c }
        \toprule
         & \multicolumn{3}{c}{Heuristics} & \multicolumn{1}{c}{RL} \\
        \cline{2-5}
         & Random & Para$_{max}$ & Para$_{min}$ & PAC-ML \\
        \midrule
        
        $\beta_{A}$ & $0.517^{+0.015}_{-0.015}$ & $0.262^{+0.002}_{-0.003}$ & $0.309^{+0.014}_{-0.015}$ & $\boldsymbol{0.203^{+0.007}_{-0.009}}$ \\
        
        $\beta_{B}$ & $0.601^{+0.007}_{-0.008}$ & $0.263^{+0.006}_{-0.004}$ & $0.396^{+0.006}_{-0.003}$ & $\boldsymbol{0.258^{+0.007}_{-0.003}}$ \\
        
        $\beta_{C}$ & $0.505^{+0.016}_{-0.012}$ & $0.267^{+0.004}_{-0.006}$ & $0.307^{+0.015}_{-0.012}$ & $\boldsymbol{0.117^{+0.003}_{-0.003}}$ \\
        
        $\beta_{D}$ & $0.465^{+0.004}_{-0.006}$ & $0.263^{+0.006}_{-0.004}$ & $0.142^{+0.027}_{-0.046}$ & $\boldsymbol{0.099^{+0.008}_{-0.007}}$ \\
        
        \bottomrule
    \end{tabular}
    \caption{Blocking rate performance of the partitioning agents on the four $\beta$ distributions (best in \textbf{bold}). Results are given as the mean across \num{3} seeds, and error bars denote the corresponding min-max confidence intervals.}
    \label{tab:blocking_rate_performance}
\end{table}

\secInLine{Partitioner baselines}
We considered three heuristic baseline partitioning strategies. $(1)$ Most prior works partition a given job across as many workers as are available up to a pre-defined environment maximum partition degree \citep{khani2021sipml, wang2022topoopt}. We refer to this strategy as `$\text{Para}_{max}$'. $(2)$ Given the low network overhead (see Figure \ref{fig:ramp_vs_meta_network_overhead}) and contentionless nature of RAMP, and given the operations' linear split-compute time dependency of our environment, a reasonable estimate for the completion time of a job with sequential run time $\text{JCT}^{\text{seq}}$ distributed across $u^{t}$ workers would be $\text{JCT} \approx \frac{\text{JCT}^{\text{seq}}}{u^{t}}$. Therefore, in light of our objective to minimise the user-defined blocking rate, we introduce a new partitioning strategy, `$\text{Para}_{min}$', which partitions the job up to the estimated minimum amount of parallelisation needed to satisfy the job's completion time requirements, $u^{t} = \lceil \frac{1}{\beta} \rceil$ (i.e. the estimated speed-up factor needed). $(3)$ For completeness, we also ran a `Random' partitioning baseline, which selects a partitioning degree randomly from amongst the number of available workers.

\secInLine{Metrics recorded}
To measure the performance of our partitioning agents, we recorded the following key metrics. $(1)$ \textit{User-defined blocking rate} (which we abbreviate to `blocking rate'): The fraction of arrived jobs which had their completion time requirements met by the cluster. $(2)$ \textit{Offered throughput}: The total `information size' of the original jobs (i.e. before partitioning was applied) processed per unit time. 
Since the open-access PipeDream job profiles used in our experiments did not contain per-operation flop/s (computational load) information, we summed the jobs' operation and dependency sizes (measured in bytes (B)) to get the total `information size' of each job. The \textit{load rate} could then be defined as the rate of job information arriving at the cluster per unit time, and the corresponding offered throughput as the rate at which this total job information was processed by the cluster. 
For a full list of metric definitions, refer to Appendix \ref{sec:metric_definitions}.

\begin{figure}[!tp]
    \centering
    \includegraphics[width=0.9\textwidth]{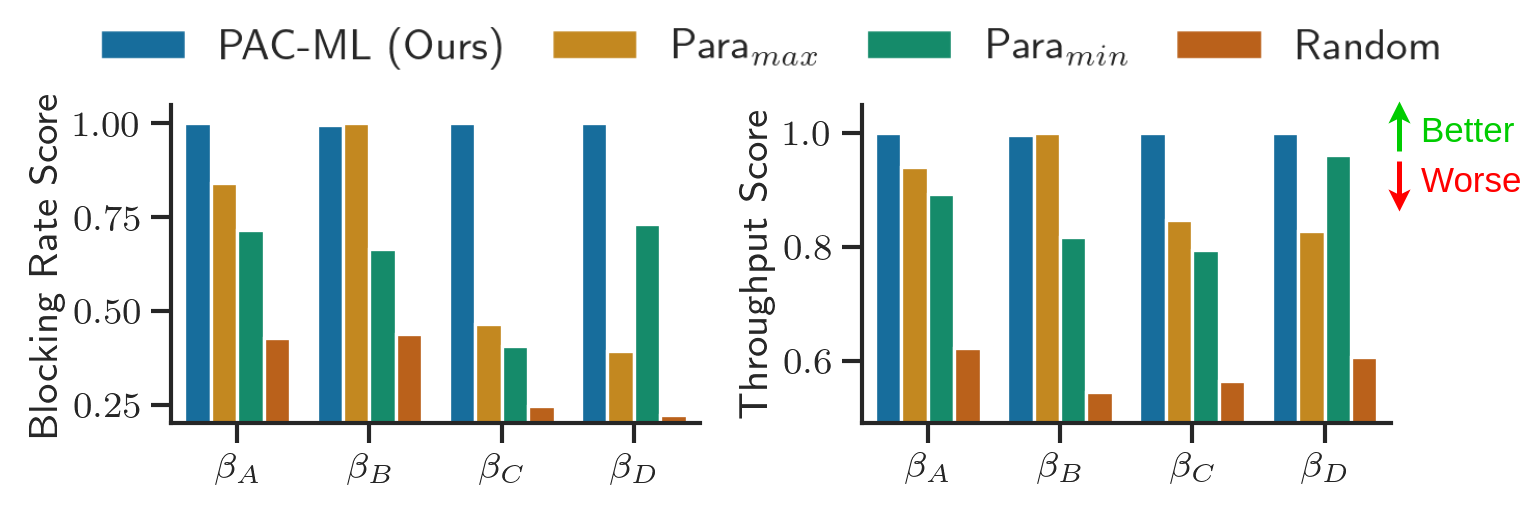}
    \caption{Validation performances (higher is better) of each partitioning agent evaluated across three seeds normalised with respect to the best-performing partitioner in each $B_{X}$ environment.}
    \label{fig:validation_bar_charts_normalised}
\end{figure}

\section{PAC-ML Partitioning Results \& Discussion}

\subsection{Performance of the PAC-ML Partitioner}

\secInLine{Comparison to the baseline partitioners}
To test the performance of each partitioning agent under different completion time requirement settings, we ran our experiments across four different $\beta$ distributions (see Section \ref{sec:experimental_setup}). We visualise the relative blocking rate and throughput performance differences between the agents in Figure \ref{fig:validation_bar_charts_normalised}, where an agent's `score' is its normalised performance relative to the best-performing agent with respect to a given metric. We evaluate these scores as $\text{score}_{\text{blocking}} = \big( \frac{\text{best\_blocking\_rate}}{\text{blocking\_rate}} \big)$, and $\text{score}_{\text{throughput}} = \big( \frac{\text{throughput}}{\text{best\_throughput}} \big)$ for each agent (refer to Appendix \ref{sec:additional_experimental_details} for all raw metric values). As shown in Table \ref{tab:blocking_rate_performance} and Figure \ref{fig:validation_bar_charts_normalised}, our PAC-ML agent achieved the best blocking rate across all four $\beta$ distributions, beating its nearest rival by $22.5\%, 1.90\%, 56.2\%, \text{ and } 30.3\%$ for $\beta_{A, B, C, D}$ respectively.



\secInLine{Comparison amongst the baseline partitioners}
Figure \ref{fig:validation_bar_charts_normalised} visualises the performance of the best PAC-ML agents on each of the four $\beta$ distribution environments compared to the baseline heuristic performances. Interestingly, the best baseline in terms of blocking rate for $\beta_{A, B, C}$ is $\text{Para}_{max}$, but this switches to $\text{Para}_{min}$ for $\beta_{D}$. On $\beta_{B}$, PAC-ML achieved roughly equivalent performance to Para$_{max}$ by learning that, on this $\beta$ demand distribution, maximum parallelisation led to the lowest blocking rates. This shows that different partitioning strategies have varying relative performances under different cluster settings. A key advantage of PAC-ML is therefore that the question of which partitioning strategy is best for a given environment need not be addressed by sub-optimal hand-crafted heuristics or environment-specific hyperparameter tuning. Instead, we have demonstrated in Table \ref{tab:blocking_rate_performance} and Figure \ref{fig:validation_bar_charts_normalised} that PAC-ML can \textit{automatically} learn performant partitioning strategies in arbitrary environment settings.


\subsection{Analysis of the PAC-ML Partitioner}

\secInLine{Offered throughput analysis}
One risk of optimising only for the blocking rate when training the PAC-ML agent is that it maximises the number of jobs accepted by prioritising small low-information jobs at the cost of a sub-optimal offered throughput; a key metric when measuring a cluster's quality of service to users. Figure \ref{fig:validation_bar_charts_normalised} shows that the offered throughput improves with the blocking rate, with the PAC-ML agent ultimately achieving the best throughput across all four $\beta$ distributions.


\begin{figure}[!tp]
    \centering
    \includegraphics[width=0.9\textwidth]{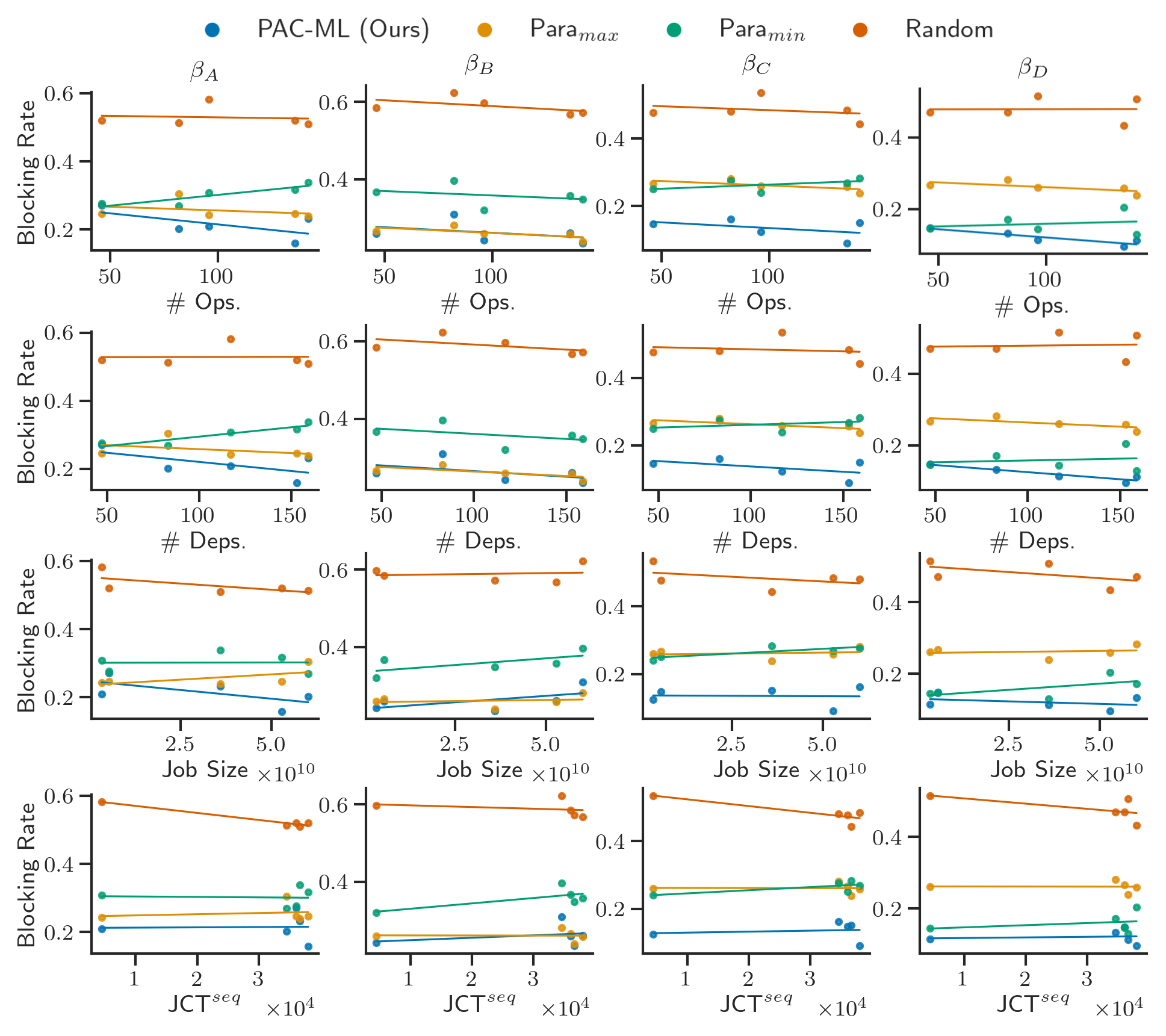}
    \caption{Mean per-job blocking rates of the five job types considered for each partitioning agent under each $\beta_{X}$ setting plotted against the number of operations (ops.), number of dependencies (deps.), the total job information size, and the sequential run time of the job were it ran on a single device (JCT$^{seq}$).}
    \label{fig:partitioner_fair_share_analysis_2}
\end{figure}

\secInLine{Bias analysis}
An important question is whether there is any bias in the kinds of jobs the PAC-ML agent learns to prioritise in order to minimise the blocking rate. To investigate this, Figure \ref{fig:partitioner_fair_share_analysis_2} shows the blocking rate vs. the original characteristics for each of the five jobs considered (see Appendix \ref{sec:job_computation_graph_data_sets} for a summary of these characteristics) for each $\beta_{X}$ distribution environment. The PAC-ML agent had little to no bias across the jobs relative to the other partitioners, with all jobs attaining approximately the same blocking rate. There was a slight bias towards the larger jobs with greater sequential completion times and more information to process, which is likely due to the fact that larger jobs occupy more resources and therefore inherently become favoured over smaller jobs.

\section{Conclusion \& Further Work}

In conclusion, we have introduced a new partitioning strategy called PAC-ML. Leveraging \ac{rl} and a \ac{gnn}, PAC-ML learns to partition computation jobs dynamically arriving at a cluster of machines such that the number of jobs which meet arbitrary user-defined completion time requirements is maximised without the need for hand-crafted heuristics or environment-dependent hyperparameter tuning. We tested our partitioner on the recently proposed RAMP optical architecture \citep{ottino2022ramp} across four distributions of user-defined completion time requirements, demonstrating up to $56.2\%$ lower blocking rates relative to the canonical maximum parallelisation strategies used by most prior works when partitioning five real deep learning jobs. We hope that our work will spur a new avenue of research into developing partitioning strategies for distributed computing. In this section, we outline potentially interesting areas of further work.

\secInLine{Exceeding completion time expectations}
In this work, we rewarded PAC-ML with $+1$ for completing a job within the user-defined maximum acceptable completion time and $-1$ for failing to do so. Although minimising the blocking rate is crucial for users, it would also be desirable to minimise the \ac{jct} as much as possible. An interesting area of further study would therefore be to incorporate this objective into the reward function, perhaps by combining the \ac{jct} speed-up factor or offered throughput with the blocking rate via multi-objective \ac{rl} \citep{hayes2022practical}.

\secInLine{Real-world experiments}
Our work has considered real open-access deep learning computation graph profiles but on a simulated optical architecture. A natural but significant next step would be to implement PAC-ML in a real distributed cluster. An important question would be whether an agent trained in a simulated environment would be capable of inferring in a real cluster at test time, or if real-world training would be needed.

\secInLine{Generalisation to unseen environments}
This study ran PAC-ML in an environment which had the same load rate, $\beta$ distribution, cluster network size, and job computation graphs at train and test time. An interesting research question would be whether PAC-ML would be able to learn on one set (or a distribution) of these parameters and then generalise to a new set at test time, or if it would need to leverage existing or new state-of-the-art methods in \ac{gnn} \citep{knyazev2019understanding, vikas2020generalization, shaohua2021generalizing} and \ac{rl} \citep{karl2019quantifying, kaixin2020improving, robert2021survey} generalisation.

\secInLine{Robustness to stochastic inter-arrival times}
In our experiments, we fixed the inter-arrival rate in order to fix the load rate. However, real clusters have variable inter-arrival times \citep{parsonson2022traffic}. Handling highly stochastic environments is a known challenge for \ac{rl} \citep{mao2019variance}, and therefore presents an interesting future research avenue for PAC-ML.

\secInLine{Combining the virtual plane}
In our work, we have considered the job partitioning task in isolation of the job placement and scheduling tasks. However, prior works have found the merging of the latter sub-tasks into a single resource management problem beneficial to performance \citep{paliwal2020reinforced}. An interesting area of further work would be to combine PAC-ML into a a single algorithm which handled job partitioning, placement, and scheduling via methods such hierarchical \ac{rl} \citep{Barto2003RecentAI, vezhnevets2017feudal, mirhoseini2018a, paliwal2020reinforced, zhang2021hierarchical} or multi-agent \ac{rl} \citep{foerster2018a}.

\section{Appendix}

\subsection{Extended Background}

\subsubsection{Parallelisation}
\label{sec:extended_background_parallelisation}

There are three main types of deep learning parallelism; \textit{data parallelism}, \textit{model parallelism}, and \textit{hybrid parallelism}.

\secInLine{Data parallelism}
Data parallelism \citep{slotnick1962solomon} is where an identical copy of the \ac{dnn} model is sent to each worker. The input training data is parallelised by sampling a training batch, splitting it into non-overlapping micro-batches, training each worker on its own micro-batch, and updating the workers' local model parameters using some method to synchronise the gradients of the parameters with respect to the training loss after each training iteration. This synchronisation step is commonly referred to as \textit{AllReduce}, and can be performed using various techniques. Data parallelism can be applied to any \ac{dnn} model regardless of its architecture, enables the use of large data sets (which are crucial for scaling model performance \citep{hoffman2022training}), and facilitates the use of large training batch sizes which can lead to smoother and faster convergence. This is a form of \textit{weak scaling}, where the \ac{jct} is decreased by reducing the total number of training iterations needed via increasing the amount of data processed per iteration as the number of workers is increased \citep{khani2021sipml}. However, it scales poorly for large models with many parameters since all parameters must fit onto a single worker and then be synchronised at the end of each training step, and has the constraint that the training data must be i.i.d. in order for parameter updates to be computed and summed across workers to attain the updated model parameters. 

\secInLine{Model parallelism}
Model parallelism \citep{karakus2021amazon} is where the \ac{dnn} model is \text{partitioned} (split) and a part of the model is sent to each worker. In the \ac{dnn} forward pass, a training batch is sampled, copied, and sent to each worker which holds layer-\num{1} of the \ac{dnn}. The layer-\num{1} worker(s) then compute the layer-\num{1} output(s) and forward them to the worker(s) which hold layer-\num{2}, and so on. In the backward pass, the gradients of the model parameters with respect to the training loss are computed by starting at the worker(s) which hold the final layer and propagating these gradients back to the layer-\num{1} workers, after which the partitioned model will be globally synchronised. Layer outputs, gradients, and activations are exchanged during the training iteration using a synchronisation step commonly referred to as \textit{AllGather}. Model parallelism facilitates the use of very large models which otherwise would not fit onto a single worker and caters for time-efficient parallelisation of computational operations where possible. This is a form of \textit{strong scaling}, where the \ac{jct} and per-worker memory utilisation are decreased via increasingly partitioning different parts the job across more workers as the number of workers is increased \citep{khani2021sipml}. However, passing gradients between workers during training can create a large communication overhead \citep{mirhoseini2017device, mirhoseini2018hierarchical}, and expert domain knowledge of the specific model architecture is needed to know how to split the model across multiple workers.

\secInLine{Hybrid parallelism}
Hybrid parallelism \citep{dean2012large} is where a combination of data and model parallelism is used to strive for the benefits of both. This can be extended to include \textit{pipeline parallelism} \citep{huang2019gpipe, narayanan2019pipedream}, where intra-batch parallelism (data and model parallelism) are combined with inter-batch parallelism (pipelining) where multiple micro-batches are processed simultaneously where possible. Hybrid parallelism can result in higher worker utilisation and the advantages of both model and data parallelism, but requires complex bidirectional pipelining across different inputs, careful model parameter versioning to ensure correct computations of the gradients during the backward pass, and each stage allocated across workers must be load-balanced to ensure roughly equivalent computational times between workers in order to maximise peak pipeline throughput.

\subsubsection{Neural Networks as Function Approximators}
\label{sec:extended_background_neural_networks}

\secInLine{Neural networks}
Neural networks are a composition of linear and non-linear (\textit{activation}) functions connected in a chain to form a \ac{dag}. Each function in the chain is a \textit{layer} parameterised by a set of weights and biases which, given enough parameters, can be trained to approximate any universal function \citep{hornik1989multilayer, montufar2014on}. Neural networks with multiple intermediary (\textit{hidden}) layers between input and output are referred to as \textit{deep} neural networks and have powerful expressivity capabilities when approximating complex non-linear functions \citep{hornik1989multilayer, montufar2014on}.

\secInLine{Graph neural networks}
Whereas standard \acp{dnn} are restricted to handling only vector- and grid-structured inputs (e.g. sentences, images, etc.), \acp{gnn} are generalised \ac{dnn} architectures which can handle graph-structured data as inputs (e.g. job \acp{dag}). Most current \acp{gnn} use the message passing paradigm by mapping each node and edge onto a vector embedding space before performing additional graph-level embeddings and readouts if desired.

Specifically, each \ac{gnn} layer usually performs four stages: (i) On each edge in the input graph, use a \textit{message function} to generate a message (representation) to pass from a source node to a set of destination nodes, where each node stores the message(s) it receives in its \textit{mailbox}; (ii) on each node in the input graph, apply an aggregate function (a vanilla reduce operation such as mean, sum, max, min, etc., or a trainable function) to the messages in its mailbox to generate an \textit{intermediate} aggregate representation of its neighbourhood; (iii) pass the intermediate aggregate representation through a trainable function to produce a \textit{final} vector embedding for each node; and (optional) (iv) if desired, at the end of the final GNN layer, pass the node embeddings through a trainable function to produce a graph-level representation. Crucially, the parameters of all message, aggregation, and forward pass functions are shared across nodes, enabling \acp{gnn} to be \textit{inductive} in that they can generalise to unseen nodes and graphs.

\subsubsection{Reinforcement Learning Algorithm}
\label{sec:extended_background_reinforcement_learning_algorithms}

Here we break down the key background components of the \ac{rl} approach used for PAC-ML.

\secInLine{Q-learning}
Q-learning \citep{watkins1989learning} is the canonical value-based algorithm which can be applied to a sequential decision making process formalised as an \ac{mdp}. It is an off-policy temporal difference algorithm whose goal is to learn the action value function mapping state-action pairs to their expected discounted future return when following a policy $\pi$; $Q^{\pi}(s, u) = \mathbb{E}_{\pi} \big[ \sum_{t^\prime=t+1}^{\infty} \gamma_{t^\prime-1} r(s_{t^\prime}) \vert s_{t} {=} s, u_{t} {=} u \big]$. By definition, an optimal policy $\pi_{*}$ will select an action which maximises the true Q-value $Q_{*}(s, u)$, $\pi_{*}(s) = \arg\max_{u^{\prime}} Q_{*}(s, u^{\prime})$. 

Concretely, the classical Q-learning algorithm maintains an action value look-up table $Q(s, u)$ mapping all possible state-action pairs to their predicted discounted return. The return is the sum of future rewards over the remainder of the episode. During training, Q-learning follows an exploration-exploitation policy. The simplest such policy is $\epsilon$-greedy, where a random action is sampled with probability $\epsilon \in [0, 1]$ and the best action, according to the current $Q$ table, is sampled with probably $1-\epsilon$. At each time step $t$, the agent in state $s_{t}$ uses this policy to select an action $u_t$ which it performs in the environment to transition to the next state $s_{t+1}$ and receive a reward $r_{t+1}$. $Q(s, u)$ is then updated according to:

\begin{equation}
    \label{eq:q_learning_update_rule}
    Q(s_{t}, u_{t}) \leftarrow Q(s_{t}, u_{t}) + \alpha \cdot \bigg( r_{t} + \gamma \cdot  \max_{u^{\prime}}Q(s_{t+1}, u^{\prime}) - Q(s_t, u_t) \bigg).
\end{equation}

On the right-hand side of Eq. \ref{eq:q_learning_update_rule}, $Q(s_t, u_t)$ is the agent's estimate of the discounted return of taking action $u_{t}$ in state $s_{t}$, $\alpha$ is the learning rate, $\gamma$ is the factor by which to discount future rewards to their present value, and $\max_{u^{\prime}}Q(s_{t+1}, u^{\prime})$ is an estimate of the future value of being in state $s_{t+1}$ and taking an `optimal' action according to $Q$. The $r_{t} + \gamma \cdot  \max_{u^{\prime}}Q(s_{t+1}, u^{\prime})$ term is called the \textit{temporal difference target}, and and the collective $r_{t} + \gamma \cdot  \max_{u^{\prime}}Q(s_{t+1}, u^{\prime}) - Q(s_t, a_t)$ term the \textit{temporal difference error}. As such, the $\max_{u^{\prime}}Q(s_{t+1}, u^{\prime})$ term treats $Q$ as an oracle from which optimal actions can be sampled. Although $Q$ is usually randomly initialised and changes at each update step, the general idea is that, with stable learning and sufficient exploration, $Q$ will converge on the true $Q_{*}$ function.

As a side note, Q-learning is a \textit{temporal difference} algorithm because, rather than using the actual returns to update $Q$ in Eq. \ref{eq:q_learning_update_rule} as done my \textit{Monte Carlo} methods, it uses a bootstrapped estimate of the future returns $\max_{u^{\prime}}Q(s_{t+1}, u^{\prime})$. Furthermore, it is an \textit{off-policy} algorithm  because the policy used to select the action $u_{t}$ at the current time step, such as $\epsilon$-greedy sampling of $Q$, is different to the policy used to select the next-state action $u^{\prime}$ when evaluating the temporal difference target, such as greedy sampling of $Q$. This is as opposed to \textit{on-policy} temporal difference algorithms, such as SARSA, which use the same action selection policy for both the current time step and for future time steps when bootstrapping.

\secInLine{Deep Q-learning}
Many practical problems have an extremely large number of possible state-action combinations. For example, the game of Go has over $10^{700}$ possible sequences; far more than the number of atoms in the universe \citep{SilverHuangEtAl16nature}. As such, modelling the action value function with a tabular approach is intractable given practical memory constraints. To enable Q-learning to be scaled to complex problems, \ac{dqn} \citep{mnih2013playing} approximates the true Q-function with a \ac{dnn} parameterised by $\theta$ such that $Q_{\theta}(s, u) \approx Q_{*}(s, u)$. 

Concretely, during training at each time step $t$, $Q_{\theta}(s, u)$ is used with an exploration strategy such as $\epsilon$-greedy to select an action and add the observed transition $T = (s_{t}, u_{t}, r_{t+1}, \gamma_{t+1}, s_{t+1})$ to a replay memory buffer \citep{Lin1992}. The network's parameters $\theta$ are then optimised with stochastic gradient descent to minimise the mean squared error loss between the \textit{online} network's predictions and a bootstrapped estimate of the Q-value,

\begin{equation}
    \label{eq:dqn_loss}
    J_{DQN}(Q) = \big[ r_{t+1} + \gamma_{t+1} \max_{u^{\prime}} Q_{\bar{\theta}}(s_{t+1}, u^{\prime}) - Q_{\theta}(s_{t}, u_{t}) \big]^{2},
\end{equation}

where $t$ is a time step uniform randomly sampled from the buffer and $Q_{\bar{\theta}}$ a \textit{target} network with parameters $\bar{\theta}$ which are periodically copied from the acting online network. The target network is not directly optimised, but is used to provide the bootstrapped Q-value estimates for the loss function. Only periodically updating the target network rather than at each learning step leads to lower variance in the bootstrapped targets at each step. This helps helps to stabilise learning and leads to better convergence \citep{mnih2013playing}.

\secInLine{Double DQN}
In the traditional Q-learning update rule of Eq. \ref{eq:q_learning_update_rule} and the DQN loss of Eq. \ref{eq:dqn_loss}, the Q-function used to select and evaluate an action for the temporal difference target is the same; $\max_{u^{\prime}}Q(s_{t+1}, u^{\prime})$ for Eq. \ref{eq:q_learning_update_rule}, and $\max_{u^{\prime}} Q_{\bar{\theta}}(s_{t+1}, u^{\prime})$ for Eq. \ref{eq:dqn_loss}. However, this can lead to an \textit{overestimation bias} where the chosen action $u^{\prime}$ is incorrectly over-valued because the same function which perceives $u^{\prime}$ as being best is also being asked to evaluate it. This can lead to high variance updates, unstable learning, and convergence on local minima. Double DQN \citep{vanhasselt2015deep} reduces overestimation by decomposing the \texttt{max} operation in the temporal difference target into \textit{action selection} and \textit{action evaluation} and performing these two tasks with two separate networks. 

Concretely, action $u^{\prime}$ is greedily selected according to the online network $Q_{\theta}$ and evaluated with the separate target network $Q_{\bar{\theta}}$. The loss term from Eq. \ref{eq:dqn_loss} then becomes:

\begin{equation}
    \label{eq:ddqn_loss}
    J_{DDQN}(Q) = \big[ r_{t+1} + \gamma_{t+1} Q_{\bar{\theta}}(s_{t+1}, \max_{u^{\prime}} Q_{\theta}(s_{t+1}, u^{\prime})) - Q_{\theta}(s_{t}, u_{t}) \big]^{2}.
\end{equation}

\secInLine{Prioritised experience replay}
Vanilla DQN replay buffers are sampled uniformly to obtain transitions for network updates. A preferable approach is to more frequently sample transitions from which there is much to learn. Prioritised experience replay \citep{schaul2016prioritized} deploys this intuition by sampling transitions with probability $p_{t}$ proportional to the last encountered absolute temporal difference error,

\begin{equation}
    p_{t} \propto | r_{t+1} + \gamma_{t+1} \max_{u^{\prime}} Q_{\bar{\theta}}(s_{t+1}, u^{\prime}) - Q_{\theta}(s_{t}, u_{t}) |^{\omega},
\end{equation}

where $\omega$ is a tuneable hyperparameter for shaping the probability distribution. New transitions are added to the replay buffer with maximum priority to ensure all experiences will be sampled at least once to have their errors evaluated.

\secInLine{n-step Q-learning}
Traditional Q-learning uses the target network's greedy action at the next step to bootstrap a Q-value estimate for the temporal difference target. Alternatively, to improve learning speeds and help with convergence \citep{sutton1998reinforcement, hessel2017rainbow}, forward-view \textit{multi-step} targets can be used \citep{sutton1998reinforcement}, where the $n$-step discounted return from state $s$ is

\begin{equation}
    r_{t}^{(n)} = \sum_{k=0}^{n-1} \gamma_{t}^{(k)} r_{t+k+1},
\end{equation}

resulting in an $n$-step DQN loss of 

\begin{equation}
    J_{DQN_{n}}(Q) = \big[ r^{(n)}_{t} + \gamma^{(n)}_{t} \max_{u^{\prime}} Q_{\bar{\theta}}(s_{t+n}, u^{\prime}) - Q_{\theta}(s_{t}, u_{t}) \big]^{2}.
\end{equation}

\secInLine{Dueling DQN}
Traditional \ac{dqn} approaches use a \ac{dnn} architecture which is not specific to \ac{rl}. Subsequently, when learning the Q-function, the entire \ac{dnn} architecture must learn to estimate the state value \textit{and} the action advantage for each action in order to learn the state-action function $Q^{\pi}(s, u)$ of being in state $s$, taking action $u$, and following policy $\pi$. However, in many problems where bootstrapped Q-learning is applied, the most important objective is to learn to estimate the value of each state rather than the effect of each action for each state. This is especially true in environments and individual states where future transitions are mainly influenced by factors other than the agent's actions. 

Leveraging the insight that in many states it is unnecessary to estimate the value of each action choice, \cite{wang2015dueling} developed a new \ac{dnn} architecture, termed `dueling DQN', which is better suited to the Q-learning task. Concretely, the dueling architecture uses the same core \ac{dnn} as standard DQN. However, rather than following the initial encoding with a single sequence of fully connected layers to get a Q-value for each possible action in the current state, dueling DQN uses two separate streams of fully connected layers. One stream, parameterised by $\beta$, estimates the state value function $V_{\theta, \beta}(s)$ (the estimated future discounted return of the current state regardless of future actions taken), and the other stream, parameterised by $\alpha$, estimates the relative action advantage function $A_{\theta, \alpha}(s, u)$ (the relative difference in the future discounted return of each action).

The outputs of the two streams are then combined via a special aggregation function to recover the state-action value function $Q$. Crucially, $V(s)$ and $A(s, u)$ must be combined into $Q(s, u)$ in such a way that they are independently identifiable from the output $Q$ values alone in order for backpropagation to be able to calculate the appropriate loss and weight updates for the separate $V(s)$ and $A(s, u)$ streams. As such, a simple $Q(s, u) = V(s) + A(s, u)$ aggregation function to get the Q-values from the two streams does not suffice. Instead, the authors tried two different aggregation schemes.

The first aggregation method subtracted the advantage of the \textit{maximum} advantage action from all advantages to make the \texttt{argmax} action's advantage $0$ and the rest $<0$,

\begin{equation}
    Q_{\theta, \alpha, \beta} = V_{\theta, \beta}(s) + \bigg( A_{\theta, \alpha}(s, u) - \max_{u^{\prime}} A_{\theta, \alpha}(s, u^{\prime}) \bigg),
\end{equation}

thus enabling $V(s)$ to be recovered at the \texttt{argmax} action's Q-value.

The second aggregation method subtracted the \textit{mean} advantage from all action advantages to centre the advantage values around $0$ (i.e. to have a mean of $0$), 

\begin{equation}
    Q_{\theta, \alpha, \beta} = V_{\theta, \beta}(s) + \bigg( A_{\theta, \alpha}(s, u) - \frac{1}{|A|} \sum_{u^{\prime}} A_{\theta, \alpha}(s, u^{\prime}) \bigg).
\end{equation}

This makes $V(s)$ recoverable from $Q(s, u)$ by estimating the $V(s)$ value which, when subtracted from each $A(s, u)$ value, leads to a set of $A(s, u)$ values which have a mean of $0$. In practice, this second approach of using the mean was found to lead to more stable learning since using a \texttt{mean} operation resulted in lower variance targets between learning steps compared to when a \texttt{max} operation was used.

As with standard Q-learning, the output of the dueling network is a set of Q-values (one for each action), therefore no change to the underlying algorithm other than a slight adjustment of the network architecture was required. By decomposing the Q-function approximator in this way, dueling \ac{dqn} is able to attain superior policy evaluation in the presence of many similar-value actions, and the authors demonstrated their architecture achieving state-of-the-art performance on the Atari $2600$ games.

\secInLine{Ape-X DQN}
Noting that state-of-the-art \ac{ml} performance is often achieved with more computation, more powerful models, and larger training data sets, \cite{horgan2018distributed} proposed Ape-X; a parallelisation approach to off-policy experience replay \ac{rl}. Concretely, rather than using a single actor-learner setup, Ape-X decouples acting from learning. It distributes many actors across a set of CPU cores each with their own instance of the environment. Each actor retains a copy of a \ac{dnn} shared across actors which it uses for action selection to accumulate experiences in parallel with other actors. These experiences are then communicated to a central shared replay buffer, where a single learner mounted on a GPU uses prioritised experience replay to sample the most important experiences for learning. Learner sampling, gradient computation, and network updates are done asynchronously with one another on separate threads, as are the periodic updates made to the actors' networks with the latest shared learner network. By using multiple actors in parallel, not only can orders of magnitude more transition data be attained for learning, but also a broader diversity of experiences can be collected by allocating a different exploration strategy to each actor and thereby avoid local optima in difficult exploration and large state-action space settings. For $N_{actors}$ distributed actors, \cite{horgan2018distributed} used a per-actor $\epsilon$-greedy exploration strategy whereby each actor $i$ had a fixed exploration probability $\epsilon_{i} = \epsilon^{1 + \frac{i}{N_{actors}-1}\cdot\alpha}$ where $\epsilon=0.4$ and $\alpha=0.7$. The authors demonstrated their approach achieving new state-of-the-art results on Atari in a fraction of the training time of prior works.

\subsection{Metric Definitions}
\label{sec:metric_definitions}












Table \ref{tab:metric_definitions} summarises the metric jargon used throughout our manuscript.

\begin{table*}[!htp]
    \centering
    \small
    \begin{tabular}{l | p{6cm}}
    \toprule
         Metric & Description \\
         
         \midrule
         
         Job completion time & Time between job arriving and being completed. \\
         
         Sequential job completion time & Time it would take to complete a job were its operations ran sequentially on a single device. \\
         
         Maximum acceptable job completion time & Maximum time allowed to complete a job. \\
         
         Speed-up factor & Factor difference between sequential job completion time and actual job completion time. \\
         
         Network overhead & Fraction of the job completion time spent communicating information between workers when no computation was taking place.  \\
         
         Blocking rate & Fraction of the arrived jobs which were successfully serviced across a given period of time. \\
         
         Job information size & Summed sizes (in bytes) of a job's operations and dependencies. \\
         
         Cluster throughput & Total \textit{partitioned job} information processed per unit time by the cluster. \\
         
         Offered throughput & Total \textit{original job} information processed per unit time by the cluster. \\
         
         Load rate & Amount of job information arriving at the cluster per unit time. \\
         
         Job inter-arrival time & Time between when two jobs arrived at the cluster. \\

    \end{tabular}
    \caption{Descriptions of the various metrics referred to throughout this manuscript.}
    \label{tab:metric_definitions}
\end{table*}

\subsection{Experimental Hardware}

All environment simulations were ran on Intel Xeon ES-2660 CPUs, and all learner network training and inference was done on either a V100 or an A100 GPU.


\subsection{Additional Simulation Details}
\label{sec:additional_simulation_environment_details}

\subsubsection{Code Structure}

We built a core RAMP simulation environment which followed a Gym-like interface \citep{brockman2016openai} but without inheriting from a Gym environment object to allow additional flexibility. We then built a wrapper `job partitioning' environment which did conform to the Gym interface but used our core RAMP simulation environment to perform the internal RAMP simulation logic. Our code base is publicly available at \url{https://github.com/cwfparsonson/ddls} for further practical implementation details.

\subsubsection{Job Allocation Procedure}
When a job arrives at the cluster, our environment uses the following ordered sequence of task executions to allocate the job:

\begin{enumerate}
    \item \textbf{Op. partitioning}: Partition the job \ac{dag}'s operations to attain a `partitioned' job \ac{dag}. 
    
    \item \textbf{Op. placement}: Place the operations in the partitioned job \ac{dag} onto a sub-set of cluster workers.
    
    \item \textbf{Op. scheduling}: For each worker, schedule the priority of its placed operations to resolve conflicts where $\geq2$ operations are ready to be executed at the same time.
    
    \item \textbf{Dep. placement}: Given the placed operations and the data dependencies which must be exchanged between operations, place the dependencies onto cluster communication links.
    
    \item \textbf{Dep. scheduling}: For each communication link, schedule the priority of its placed dependencies to resolve conflicts where $\geq2$ dependencies are ready to be communicated at the same time.

\end{enumerate}

\subsubsection{Job Allocation Methods}

Each of the above allocation procedure tasks can be performed by any algorithm, heuristic, or learning agent. In our work, we use the following methods:

\begin{enumerate}
    \item \textbf{Op. partitioning}: PAC-ML, Para$_{max}$, Para$_{min}$, or Random. See the main manuscript for details.
    
    \item \textbf{Op. placement}: A first-fit heuristic customised for the requirements of RAMP. See Section \ref{sec:first_fit_operation_placement_in_ramp} below for details.
    
    \item \textbf{Op. scheduling}: Shortest remaining processing time \citep{Cai2016, Alizadeh2013, Hong2012}. Given a set of operations placed on a worker, the operation with the shortest remaining run time will have the highest priority and therefore be executed first wherever two operations on the same worker request to be executed at the same time.  
    
    \item \textbf{Dep. placement}: Shortest path \& first-fit. Given a set of operation placements, for any dependencies which need to be transferred through the network (i.e. for dependencies with size $>0$ and whose parent operation is placed on a separate worker from the child operation), $(1)$ first-fit select a path from the $k-$shortest path with available light channel(s), and $(2)$ first-fit select an available channel.
    
    \item \textbf{Dep. scheduling}: Shortest remaining processing time. Given a set of dependencies placed on a communication link channel, the dependency with the shortest remaining processing time (i.e. the lowest amount of information left to be transferred) will have the highest priority and therefore be communicated first wherever two dependencies on the same link channel request to be transported at the same time.

\end{enumerate}

\subsubsection{First-Fit Operation Placement in RAMP}
\label{sec:first_fit_operation_placement_in_ramp}

The original RAMP paper of \cite{ottino2022ramp} did not specify an operation placement heuristic which conformed to the RAMP placement rules (see Section \ref{sec:background}). Here, we propose a simple first-fit heuristic which conforms to these rules whilst making the placement problem tractable for large cluster networks.

The basic idea behind partitioning and placement in the scenario described in this work is to exploit the network efficiencies of RAMP as much as possible. In particular, this means maximising the use of RAMP's highly efficient collective operations. For a generic partitioned \ac{dag}, in the backward pass, collectives happen for each operation when weights/gradients are shared between sub-operations. If both a parent and child operation are placed on the same set of (RAMP symmetry adherent) workers, then when the parent communicates its output to the child's input in the forward pass this will also constitute a collective operation. As such the placement heuristic implemented here seeks to primarily maximise the amount that these two conditions are encountered. Given some operation, $o$, that has been partitioned into $N$ equal sub-operations, $o_i$ and needs to be placed, the placement is handled as:

\begin{enumerate}
    \item If a parent of $o$ has been partitioned and placed across $N$ servers which adhere to the RAMP symmetry conditions, and if these servers each have enough memory to store $o_i$, then place $o$ across this set of $N$ servers. This ensures collective operations can happen in both the forward and backward pass.
    
    \item Otherwise, check if a set of $N$ workers can be found in the network that adheres to the RAMP symmetry requirements. This is achieved by sliding the various possible symmetric shapes over the topology until a suitable one (or none) is found. This ensures collective operations in the backward pass only.
\end{enumerate}

Allocating in this way ensures that every partitioned operation can exploit RAMP's efficient collective operation process on the backward pass, and where possible can also exploit it on the forward pass when receiving information from (one of) its parents.

\subsubsection{Evaluating the job completion time}
The time to complete each operation was taken from the real computation job profiles of the \ac{dnn} jobs considered (see Section \ref{sec:job_computation_graph_data_sets}). To calculate the communication time of point-to-point information transfers and of the MPI collectives, we used the equations and code of \cite{ottino2022ramp}.

\subsubsection{Possible Causes of a Job Being Blocked}
A job is blocked when either $\text{JCT} > \beta \cdot \text{JCT}^{\text{seq}}$ (i.e. failing to meet user's chosen JCT requirement) or when the cluster does not have enough available resources to service the job. The possible causes of this latter form of blocking are:

\begin{itemize}

    \item Prior jobs using up too many cluster resources when later jobs arrive;
    
    \item the minimum operation run time quantum not being low enough to partition the operations enough times to lead to the desired JCT;
    
    \item mounted worker operation scheduling conflicts for partitioned operations mounted on the same worker leading to longer run times, since one worker can only execute one operation at a time; and
    
    \item excessive communication overheads incurring from over-partitioning of the job.
    
\end{itemize}

\subsection{Job Computation Graph Data Sets}
\label{sec:job_computation_graph_data_sets}

\begin{figure*}[!tp]
    \centering
    \includegraphics[width=0.99\textwidth]{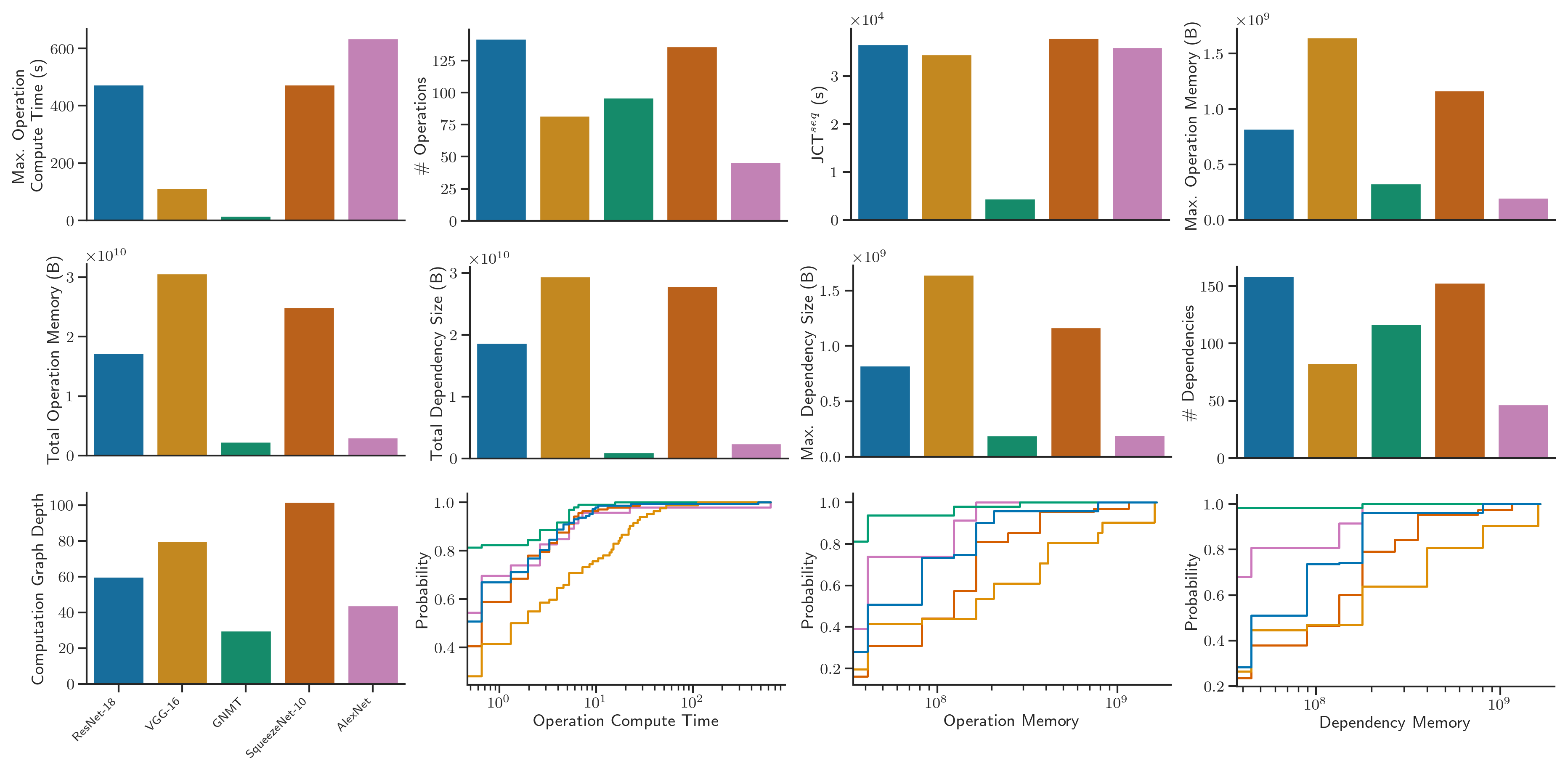}
    \caption{Visualisation of the characteristics of the deep learning computation graphs used for our experiments before partitioning. The bottom left sub-figure contains the model colour code scheme for all other sub-figures. The statistics shown are for the operations and dependencies which need to be executed and satisfied to conduct one training iteration. Therefore, to carry out $N_{iter}$ training steps, the computation graph would need to be executed $N_{iter}$ times. Computation time units are reported in seconds, and memory units in bytes.}
    \label{fig:pipeline_computation_graph_characteristics}
\end{figure*}

\begin{figure}[!tp]
    \centering
    \includegraphics[width=0.6\textwidth]{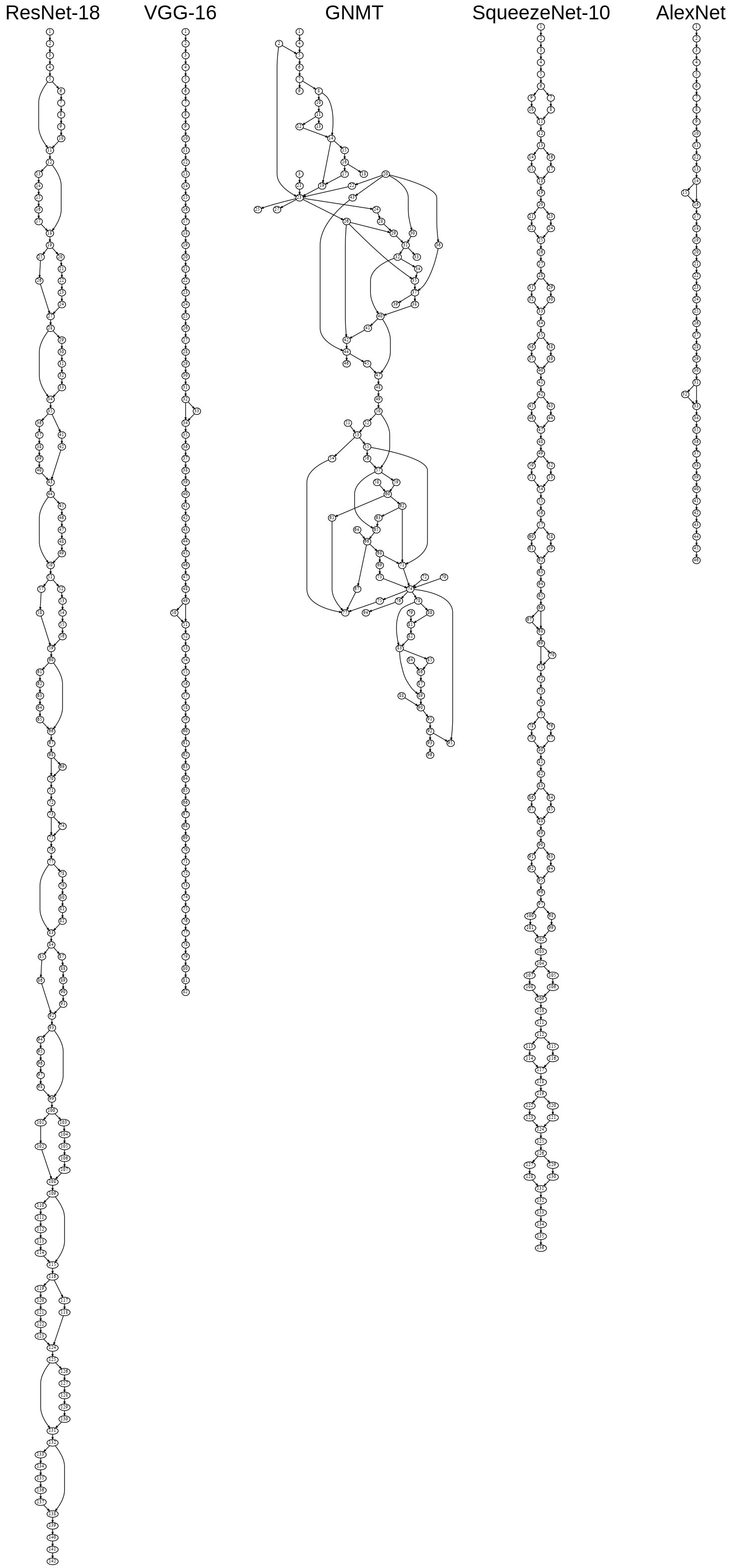}
    \caption{Deep learning computation graphs used for our experiments before partitioning. Each computation graph represents the operations and dependencies which need to be executed and satisfied to conduct one forward and one backward pass through the neural network. Therefore, to carry out $N_{iter}$ training steps, the computation graph would need to be executed $N_{iter}$ times.}
    \label{fig:pipedream_computation_graphs}
\end{figure}

All computation graphs used in our experiments were taken from the open-access PipeDream computation graph data set \citep{narayanan2019pipedream}. Figure \ref{fig:pipeline_computation_graph_characteristics} shows a visualisation of the key computation graph characteristics for each neural network model considered, where the numbers reported are for one training iteration (i.e. one forward and backward pass through the model). Table \ref{tab:pipeline_computation_graph_characteristics} reports the same characteristics but in tabular form. Finally, for completeness, Figure \ref{fig:pipedream_computation_graphs} shows the actual job \acp{dag} of the models used.

\begin{sidewaystable}[!htp]
    \centering
    \scriptsize
    \begin{tabular}{l r r r r r r r r r r c c c c c c c c c c} 
    \toprule
         Model & \# ops. & $\text{JCT}^{\text{seq}}$ & Max. op. comp. time & $\Sigma$ op. mem. & Max. op. mem. & Depth & \# deps. & $\Sigma$ dep. size & Max. dep. size \\
         
         \midrule
         
         ResNet-18 & \num{142} & \num{36668.35} & 473.625 & \num{17.25866e9} & \num{0.8221212e9} & \num{60} & \num{159} &	\num{18.73329e9} & \num{0.8220836e9} \\ 
         VGG-16 & \num{82} & \num{34525.35} & \num{113.330} & \num{30.62530e9} & \num{1.644315e9} & \num{80} & \num{83} & \num{29.46706e9} & \num{1.644167e9} \\
         
         GNMT & \num{96} & \num{4470.80} & \num{15.88} & \num{2.368447e9} & \num{3.269491e8} & \num{30} & \num{117} & \num{1.027801e9} & \num{0.1944371e9} \\
         
         SqueezeNet-10 & \num{136} & \num{38000.15} & \num{474.637} & \num{24.96262e9} & \num{1.168007e9} & \num{102} & \num{153} & \num{27.91009e9} & \num{1.167950e9} \\
         
         AlexNet & \num{46} & \num{36061.15} & \num{635.902} & \num{3.046234e9} & \num{0.1983396e9} & \num{44} & \num{47} & \num{2.422161e9} & \num{0.1982464e9}

    \end{tabular}
    \caption{Summary of the characteristics of the deep learning computation graphs used for our experiments before partitioning. The statistics shown are for the operations (`ops.') and dependencies (`deps.') which need to be executed and satisfied to conduct one training iteration. Therefore, to carry out $N_{iter}$ training steps, the computation graph would need to be executed $N_{iter}$ times. Computation (`comp.') time units are reported in seconds, and memory (`mem.') units in bytes.}
    \label{tab:pipeline_computation_graph_characteristics}
\end{sidewaystable}

\subsection{Neural Network Architecture}
\label{sec:neural_network_architecture}



\begin{figure}[!tp]
    \centering
    \includegraphics[width=0.98\textwidth]{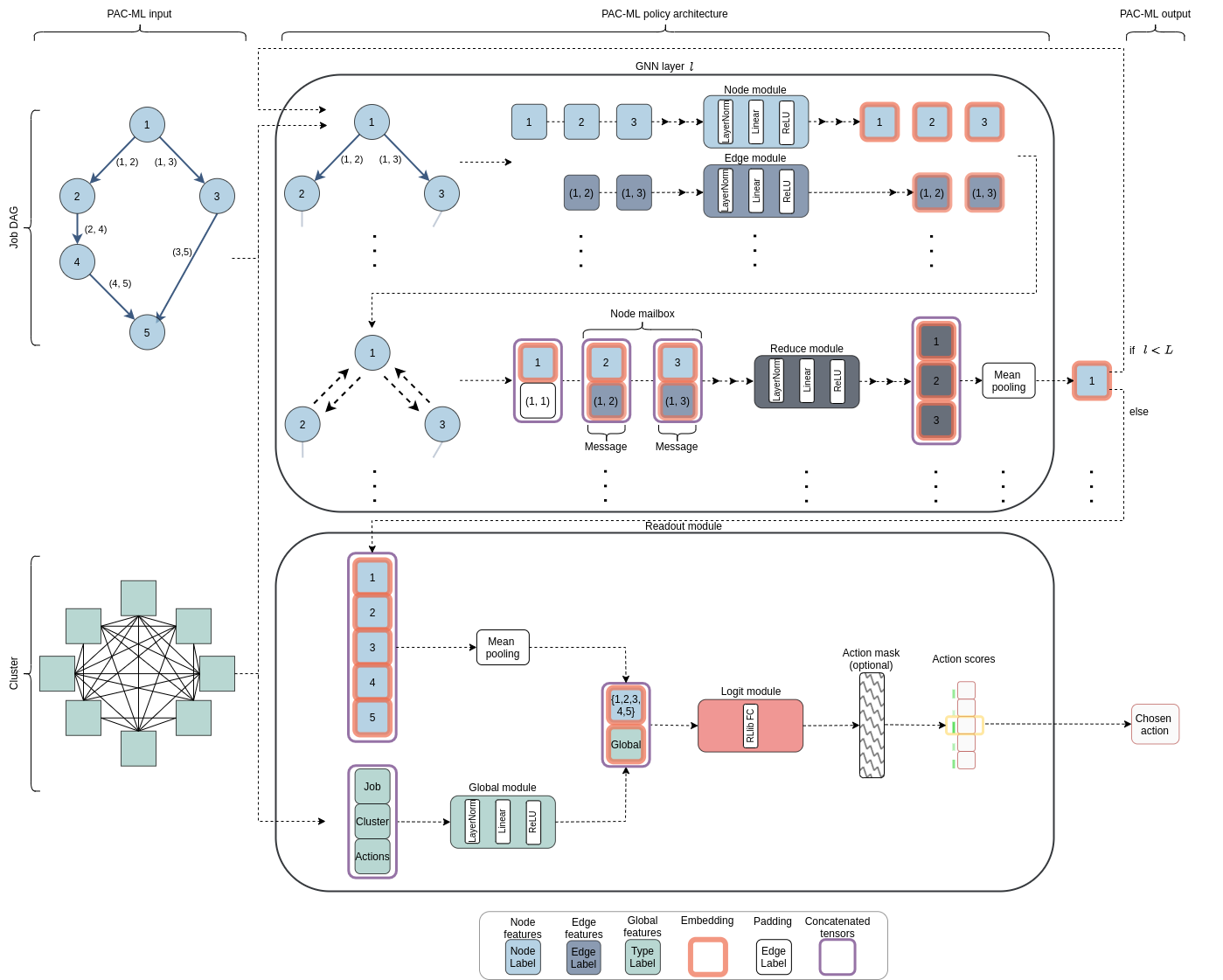}
    \caption{Schematic of the \ac{dnn} architecture with $|L|$ \ac{gnn} layers used to parameterise the policy of PAC-ML. The \ac{gnn} is similar to that of GraphSAGE with mean pooling \citep{Hamilton2017inductive}. Each GNN layer $l \in L$ contains a node, edge, and reduce \ac{dnn} module and ultimately learns to create an embedded representation for each node in a given job DAG. These per-node embeddings are then passed, along with any global job, cluster, and action features, to a readout module. The readout module ultimately generates scores for each possible action, which enables an action to be selected following a given exploration-exploitation policy being followed. For clarity, this figure only shows the GNN embedding-generation process for node $1$. See accompanying text for a detailed explanation of this architecture and the accompanying figure.}
    \label{fig:pac_ml_nn_architecture}
\end{figure}

As shown in Fig. \ref{fig:pac_ml_nn_architecture}, we used a message passing \ac{gnn} similar to GraphSAGE with mean pooling \citep{Hamilton2017inductive} to parameterise the PAC-ML policy. Table \ref{tab:dnn_hparams} summarises the hyperparameters used for the components of this \ac{dnn}. We note that we did not perform extensive hyperparamter tuning on the GNN architecture. Below is a detailed explanation of this architecture.

\secInLine{GNN}
First, the GNN layer takes in the DAG's node and edge features and generates an embedding for each node and edge in the graph. Then, each local node's nearest neighbour ($1$-hop away) sends the local node a message (`message passing') which is the neighbouring nodes' embeddings concatenated with their connected edges' embeddings. These messages are stored in the local node's `mailbox', which now contains information about the node's neighbourhood. To ensure consistent dimensioning with the received messages, a dummy zero-padded edge embedding is concatenated with the local node's embedding. Next, the reduce module takes the local and message embeddings and generates a reduced representation for each. Finally, to generate a layer-$l$ output embedding for the local node, the element-wise mean of the reduced embeddings is taken (`mean pooling'). Note that this embedding process is done for each node in the DAG, but for clairty Fig. \ref{fig:pac_ml_nn_architecture} only follows node $1$.

If $l < L$ (i.e. if this is not the last GNN layer), these final node embeddings are used as new features for the original DAG's nodes and are passed to the next GNN layer. If $l \equiv L$, then the node embeddings are passed to the readout module. Note that $(1)$ the node, edge, and reduce modules are shared across the aforementioned operations within a given GNN layer when generating node embeddings, but not across different GNN layers, and $(2)$ the $l^{th}$-layer's output node embeddings will contain information about the node's neighbourhood from up to $l$ hops away.

\secInLine{Readout}
The readout module takes the GNN's node embeddings and the job's and cluster's global features as input. To convert the node-level embeddings of the GNN into a representation of the overall job DAG, their element-wise mean is taken. To generate an embedding capturing the global job, cluster, and action information, a global \ac{dnn} module is used. The DAG and global embeddings are then concatenated and passed to a logit module, which in turn generates a vector of (optionally masked) scores for each possible action in the environment. Finally, based on these scores and the exploration-exploitation policy being followed, an action is selected.

\begin{table*}[!htp]
    \centering
    \small
    \begin{tabular}{l | r}
    \toprule
         Parameter & Value \\
         
         \midrule

         Message passing \# hidden dimensions & \numprint{64} \\
         
         Message passing \# output dimensions & \numprint{32} \\

         Reduce module \# hidden dimensions & \numprint{64} \\

         Reduce module \# output dimensions & \numprint{64} if $l < L$, else $16$ \\

         Global module \# hidden dimensions & \numprint{8} \\

         Global module \# output dimensions & \numprint{8} \\

         Logit module RLlib FC net \# layers & \numprint{1} \\

         Logit module RLlib FC net \# hidden dimensions & \numprint{256} \\

         All modules' activation & ReLU \\

         GNN \# layers $L$ & \numprint{2} \\

         Apply action mask & False \\

    \end{tabular}
    \caption{Hyperparamters used for the PAC-ML ApeX-DQN DNN policy architecture shown in Fig. \ref{fig:pac_ml_nn_architecture}. Note that the `message passing' dimensions refer to the dimensions of the concatenated node and edge modules' embeddings, so the dimensions of these modules' hidden and output embeddings will be half the corresponding `message passing' dimension. Due to the RLlib implementation of Ape-X DQN, we did not apply an action mask, but instead included the action mask in the global features given to the model and used the reward signal to train the agent to avoid selecting invalid actions.}
    \label{tab:dnn_hparams}
\end{table*}

\subsection{Reinforcement Learning Algorithm}
\label{sec:reinforcement_learning_algorithm}

\secInLine{Approach}
Given the stochastic nature of our dynamic cluster environment setting, we hypothesised that a value-based \ac{rl} method would be best suited to our setting \citep{mao2019variance}. We did try the PPO \citep{schulman2017ppo} actor-critic method but found performance to be worse, although we leave a full analysis of alternative \ac{rl} algorithms to future work. 

As stated in the main manuscript, we used the state-of-the-art value-based Ape-X DQN \ac{rl} algorithm \citep{horgan2018distributed} to attain the PAC-ML policy. Concretely, we used the Ape-X parallelisation approach with double Q-learning action selection-evaluation \citep{vanhasselt2015deep} and multi-step bootstrapped learning targets \citep{sutton1998reinforcement, hessel2017rainbow}, prioritised experience replay \citep{schaul2016prioritized}, a dueling \ac{dqn} network architecture \citep{wang2015dueling}, and a per-actor $\epsilon$-greedy exploration algorithm. For a breakdown of each of these components, refer to Appendix \ref{sec:extended_background_reinforcement_learning_algorithms}.

\begin{figure*}[!tp]
    \centering
    \includegraphics[width=0.96\textwidth]{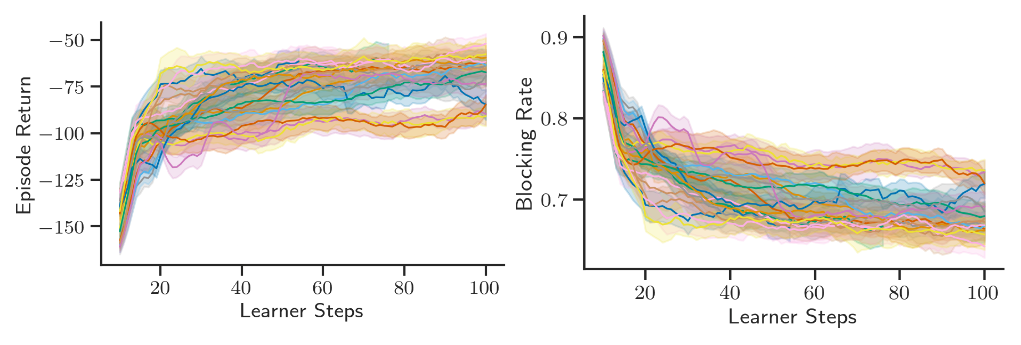}
    \caption{Validation performance of the Ape-X DQN hyperparameter sweep. Each agent was trained for $100$ learner steps, and at each learner step a validation was performed across $3$ seeds - the mean metrics with their min-max interval bands are plotted for each hyperparameter set.}
    \label{fig:learner_hparam_tuning_validation_curve}
\end{figure*}

\secInLine{Hyperparameters}
To select the algorithm hyperparameters, we conducted a Bayesian search across the search space summarised in Table \ref{tab:apex_dqn_training_parameters}, with simulations conducted in a light 32-worker RAMP environment with a maximum simulation run time of \num{2e5} seconds to speed up the search. We adopted similar search ranges to those used by \cite{kurach2019google, hoffman2020acme, parsonson2022retro}. For each set of hyperparameters, we ran the algorithm for $100$ learner steps (a.k.a. training epochs), and performed a validation across $3$ seeds at each learner step (see Figure \ref{fig:learner_hparam_tuning_validation_curve}). We selected the parameter set with the highest episode return across the $3$ seeds (see Table \ref{tab:apex_dqn_training_parameters}). We also report the importance of each parameter with respect to the total episode return. The importance is calculated by training a random forest with all algorithm hyperparameters as inputs and the episode return as the target output, with the per-feature (hyperparameter) importance values predicted by random forest reported accordingly \citep{fabros2018wiki, howard2018intro}. All our experiments used the same per-actor $\epsilon$-greedy exploration as \cite{horgan2018distributed}.

\begin{sidewaystable}[!htp]
    \centering
    \footnotesize
    \begin{tabular}{l r r r c c c c} 
    \toprule
         Parameter & Search Range & Best Value & Importance  \\
         
         \midrule

         Discount factor $\gamma$ & $\{ 0.99, 0.993, 0.997, 0.999, 0.9999 \}$ & \num{0.999} & \num{0.004} \\ 
         Learning rate & Log-uniform values ( \num{1e-7}, \num{1e-3} ) & \num{4.121e-7} & \num{0.045} \\ 
         $v_{min}$ & $\{ -1, -10, -100, -200, -1000 \}$ & \num{-1000} & \num{0.01} \\
         $v_{max}$ & $\{ 1, 10, 100, 200, 1000 \}$ & \num{1000} & \num{0.004} \\
         Target network update frequency & $\{$ \num{1e3}, \num{1e4}, \num{1e5} $\}$ & \num{1e5} & \num{0.001} \\
         Prioritised replay $\alpha$ & $\{ 0.1, 0.4, 0.5, 0.6, 0.7, 0.8, 0.9 \}$ & \num{0.9} & \num{0.04} \\
         Prioritised replay $\beta$ & $\{ 0.1, 0.4, 0.5, 0.6, 0.7, 0.8, 0.9 \}$ & \num{0.1} & \num{0.047} \\
         $n$-step & $\{ 1, 3, 5, 10 \}$ & \num{3} & \num{0.227} \\
         $\#$ CPU workers & \num{32} & \num{32} & $-$ \\
         $\#$ GPU workers & \num{1} & \num{1} & $-$ \\
         Batch mode & Truncated episodes & Truncated episodes & $-$ \\
         Rollout length & \num{50} & \num{50} & $-$ \\
         Train batch size & \num{512} & \num{512} & $-$ \\
         Optimiser & Adam & Adam & $-$ \\
         Dueling & True & True & $-$ \\
         $\#$ atoms & \num{1} & \num{1} & $-$ \\
         Noisy & False & False & $-$ \\
         Double $Q$ & True & True & $-$ \\ 
         Replay buffer capacity & \num{100000} & \num{100000} & $-$ \\
         Learning starts & \num{10000} & \num{10000} & $-$ \\
         Prioritised replay TD-error $\epsilon$ & \num{1e-6} & \num{1e-6} & $-$ \\

    \end{tabular}
    \caption{Ape-X DQN training parameter sweep search range, best value found, and corresponding parameter importance.}
    \label{tab:apex_dqn_training_parameters}
\end{sidewaystable}

We note that our \ac{rl} algorithms were implemented using the open-source RLlib library \citep{liang2018rllib} and hyperparameter tuning was done using Weights \& Biases \citep{wandb2020experiment}.

\subsubsection{Final Learning Curves}

\begin{figure}[!tp]
    \centering
    \includegraphics[width=0.9\textwidth]{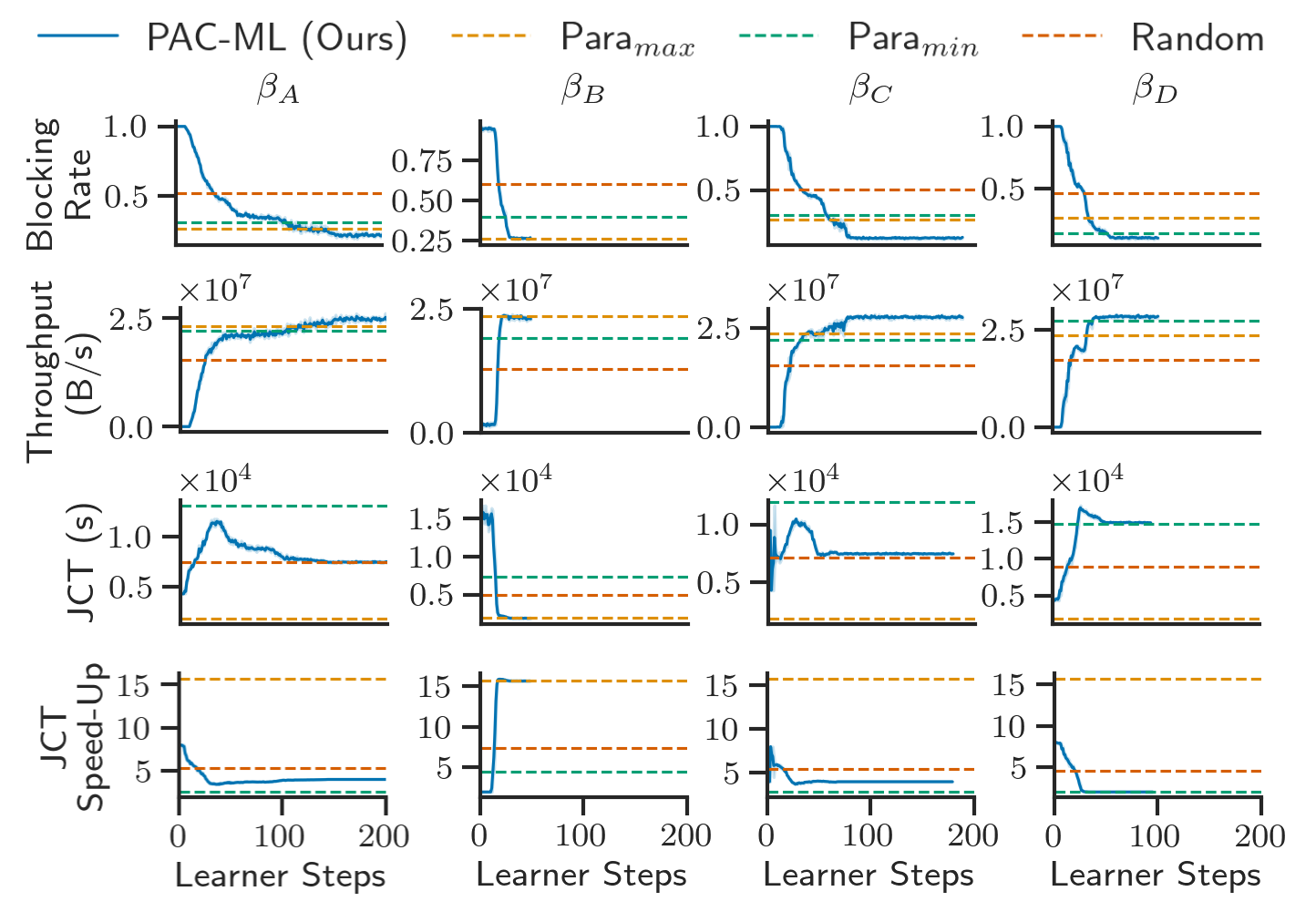}
    \caption{Validation curves of the PAC-ML agent trained in four different $\beta$ distribution environments. At each learner step (update to the GNN), the agent was evaluated across 3 seeds, with the mean blocking rate, offered throughput, \ac{jct}, and \ac{jct} speed-up (relative to the jobs' sequential run time $\text{JCT}^{\text{seq}}$) performance metrics reported as well as their min-max confidence intervals. For reference, the performances of the baseline heuristic partitioners are also plotted.}
    \label{fig:validation_curves_initial_demo}
\end{figure}

For completeness, Figure \ref{fig:validation_curves_initial_demo} shows the learning curves of the tuned PAC-ML agents in each $\beta_{X}$ environment superimposed on the baseline agents' performances. At each learner step, the PAC-ML agent was evaluated across three seeds in the validation environment.

\subsection{Additional Experimental Results}
\label{sec:additional_experimental_details}

\begin{figure}[!tp]
    \centering
    \includegraphics[width=0.9\textwidth]{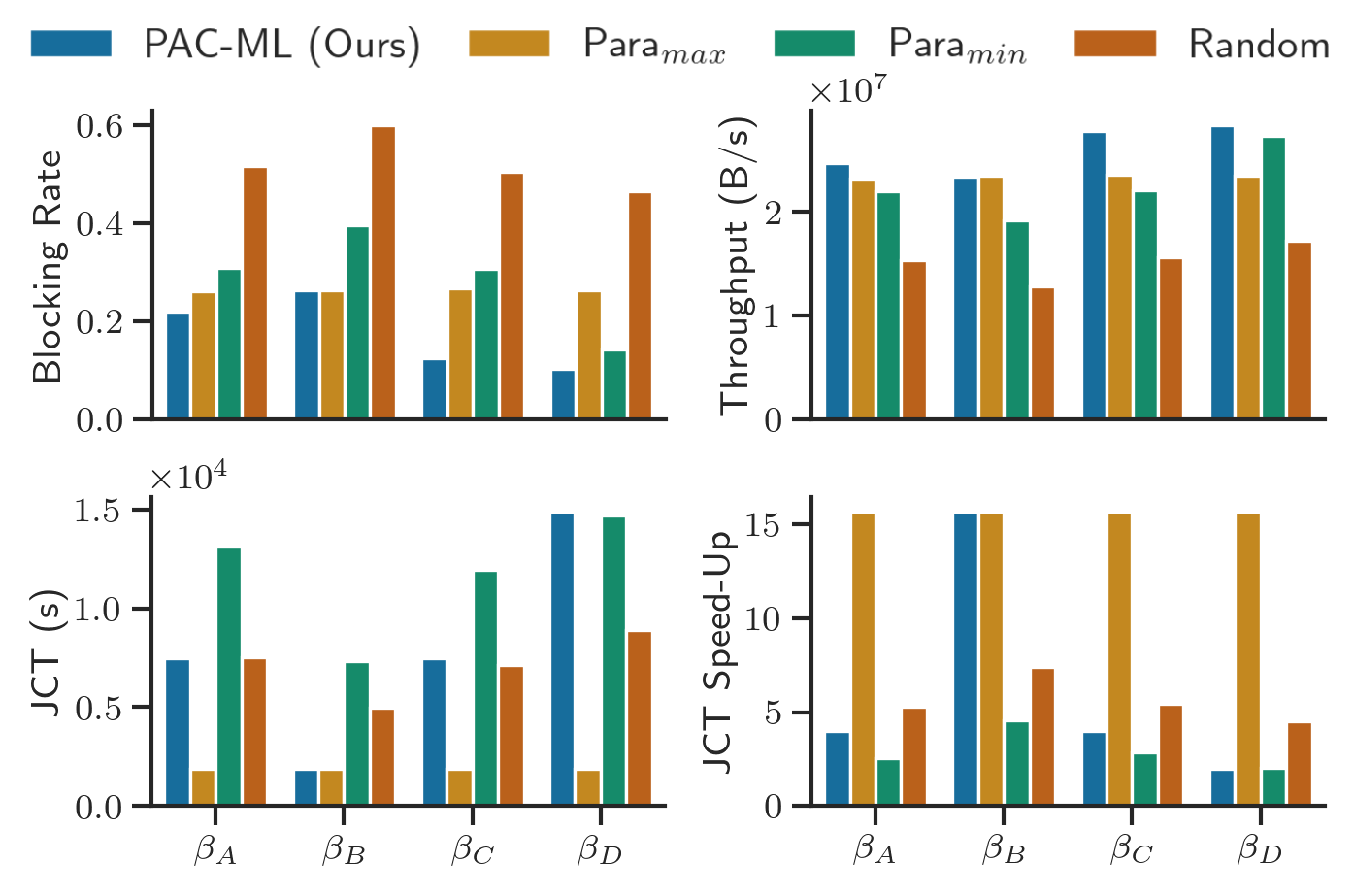}
    \caption{Validation performances of each partitioning agent evaluated across three seeds, with the mean blocking rate, offered throughput, \ac{jct}, and \ac{jct} speed-up (relative to the jobs' sequential run time $\text{JCT}^{\text{seq}}$) performance metrics reported.}
    \label{fig:validation_bar_charts}
\end{figure}

Figure \ref{fig:validation_bar_charts} shows the performance of the agents in terms of raw blocking rate, throughput, \ac{jct}, and \ac{jct} speed-up.

\section{Funding and Acknowledgments}

\section*{Funding}EPSRC Distributed Quantum Computing and Applications EP/W032643/1; the Innovate UK Project on Quantum Data Centres and the Future 10004793; OptoCloud EP/T026081/1; TRANSNET EP/R035342/1; the Engineering and Physical Sciences Research Council EP/R041792/1 and EP/L015455/1; the Alan Turing Institute; and Horizon Europe Dynamos.









\bibliographystyle{elsarticle-harv} 
\bibliography{main}

\end{document}